\documentclass[10pt,twocolumn,twoside]{IEEEtran}
\renewcommand{\baselinestretch}{0.96}

\usepackage{tcolorbox}
\usepackage{graphicx}
\usepackage{xcolor}
\usepackage{theorem}
\usepackage{times,amsmath,epsfig}
\usepackage{amssymb}
\usepackage{subfigure}
\usepackage{cite}
\usepackage{url}
\usepackage{multicol}
\usepackage{bm}
\usepackage{mathtools}
\usepackage[normalem]{ulem}

\usepackage{algorithmic}
\usepackage{algorithm}
\usepackage{tikz}
\usetikzlibrary{shapes,arrows}
\input{mysymbol.sty}
\usepackage[colorinlistoftodos]{todonotes}

\newtheorem{myexample}{\bf Example}

\newtheorem{mylemma}{\bf Lemma}
\newtheorem{myproposition}{\bf Proposition}
\newtheorem{remark}{\bf Remark}

\DeclareMathOperator{\grad}{grad}

\title{Gradient-Based Spectral Embeddings of\\ Random Dot Product Graphs}

\author{\IEEEauthorblockN{Marcelo Fiori, Bernardo Marenco, Federico Larroca, Paola Bermolen, and Gonzalo Mateos}
	\thanks{Work in this paper is supported in part by CSIC (I+D project 22520220100076UD) and the NSF awards CCF-1750428, CCF-1934962. Part of the results in this paper appeared at the \textit{2022 EUSIPCO} and \textit{Asilomar Conferences}~\cite{RDPG_EUSIPCO_22,RDPG_Asilomar_22}. \emph{(Corresponding author: Gonzalo Mateos)}.
		
		Marcelo Fiori, Bernardo Marenco, Federico Larroca, and Paola Bermolen are with the Facultad de Ingenier\'ia, Universidad de la Rep\'ublica, Montevideo 11000, Uruguay (e-mail: mfiori@fing.edu.uy; bmarenco@fing.edu.uy;
		flarroca@fing.edu.uy; paola@fing.edu.uy). Fiori, Marenco and Bermolen are also with the Centro Interdisciplinario en Ciencia de Datos y Aprendizaje Automático (CICADA), Universidad de la República, Uruguay.
		
		Gonzalo Mateos is with the Department of Electrical and Computer Engineering, University of Rochester, Rochester, NY 14627 USA (e-mail: gmateosb@ece.rochester.edu).}}

\begin{document}
	\maketitle

	\begin{abstract}%
		The Random Dot Product Graph (RDPG) is a generative model for relational data, where nodes are represented via latent vectors in low-dimensional Euclidean space. RDPGs crucially postulate that edge formation probabilities are given by the dot product of the corresponding latent positions. Accordingly, the \emph{embedding} task of estimating these vectors from an observed graph is typically posed as a low-rank matrix factorization problem. The workhorse Adjacency Spectral Embedding (ASE) enjoys solid statistical properties, but it is formally solving a surrogate problem and can be computationally intensive. In this paper, we bring to bear recent advances in non-convex optimization and demonstrate their impact to RDPG inference. We advocate first-order gradient descent methods to better solve the embedding problem, and to organically accommodate broader network embedding applications of practical relevance. Notably, we argue that RDPG embeddings of directed graphs loose interpretability unless the factor matrices are constrained to have orthogonal columns. We thus develop a novel feasible optimization method in the resulting manifold. The effectiveness of the graph representation learning framework is demonstrated on reproducible experiments with both synthetic and real network data. Our open-source algorithm implementations are scalable, and unlike the ASE they are robust to missing edge data and can track slowly-varying latent positions from streaming graphs. 
	\end{abstract}
	
	
	%
	\section{Introduction}\label{sec:intro}
	
	\IEEEPARstart{D}{uring} the last few years the Random Dot Product Graph (RDPG) model has emerged as a popular generative model for random graphs~\cite{priebe2018survey}. This latent position model associates each node $i\in\{1,\ldots,N\}$ with a vector $\bbx_i\in \mathcal{X} \subset \reals^d$; a so-termed node embedding where typically $d\ll N$. In its simplest rendition for undirected and unweighted graphs without self loops, RDPGs specify an edge exists between nodes $i$ and $j$ with probability given by the inner-product of the corresponding embeddings; independently of all other edges. That is to say, entry $A_{ij}$ of the random adjacency matrix $\bbA\in \reals^{N\times N}$ has the Bernoulli($\bbx_i^\top \bbx_j$) distribution. 
	
	This apparent simplicity does not compromise expressive power. Indeed, one can verify that Erd\"os-R\'enyi (ER) graphs or Stochastic Block Models (SBMs) with a positive semi-definite (PSD) probability matrix are particular cases of an RDPG. Several other more sophisticated models may be included as particular cases of RDPG~\cite{priebe2018survey}, being this expressiveness one of the main reasons behind its popularity.
	%
	A second reason is the model's interpretability. Since the connection probability is given by the dot product of the embeddings, the affinity between the corresponding nodes is directly captured by their alignment. We may for instance rely on visual inspection of the nodes' vector representations (possibly after further dimensionality reduction if $d>3$) to screen for community structure, or carry out angle-based clustering of nodes~\cite{scheinerman2010rdpg,Lyzinski2017community}. 
	
	The restriction to undirected graphs is overcome by considering the directed version of RDPG, where each node is assigned \emph{two} vectors $\bbx_i^l,\bbx_i^r\in\mathcal{X}\subset\reals^{d}$. A directed edge from node $i$ to $j$ exists with probability $(\bbx_i^l)^\top\bbx_j^r$~\cite{priebe2018survey}. The interpretation is analogous to the undirected case, with $\bbx^l_i$ and $\bbx^r_i$ now representing the propensity of node $i$ towards establishing outgoing and accepting incoming directed edges, respectively. For extensions to weighted graphs, see e.g.,~\cite{marenco2021tsipn}. 
	
	Rather than generating random graphs from vectors, focus in this paper is on the inverse \emph{embedding problem} at the heart of graph representation learning (GRL)~\cite{hamilton2020representation}: given the adjacency matrix $\bbA$ of a graph adhering to an RDPG model, estimate the latent position vectors for each node. Because of the RDPG definition in terms of dot products, the latent position vectors are only identifiable up to a 
	common rotation. For both undirected and directed graphs (digraphs), the workhorse approach is based on a spectral decomposition of $\bbA$ -- an estimator known as Adjacency Spectral Embedding (ASE)~\cite{priebe2018survey}.  
	
	\subsection{Challenges facing the ASE}\label{subsec:ase_challenges}
	
	Although the ASE is widely adopted and its statistical properties (such as consistency and its limiting distribution as $N\to \infty$) are well documented~\cite{priebe2018survey}, it does present drawbacks which we seek to overcome.\vspace{2pt}
	
	\noindent\textbf{Large data.} The first challenge pertains to scalability. Computing the spectrum of a large adjacency matrix $\bbA$, even only the $d$ dominant components, is computationally intensive and constitutes a bottleneck of state-of-the-art ASE implementations~\cite{graspy}, especially when multiple graphs are to be embedded. Recent work explicitly comments on the difficulty of scaling spectral-based inference of RDPGs to large graph settings~\cite{gallagher2021spectral}.\vspace{2pt} 
	
	\noindent\textbf{Missing data.} A second drawback of ASE is its inability to properly account for missing data, meaning unobserved entries in $\bbA$. On a related note, the ASE neglects the all-zeros diagonal in the adjacency matrix. These limitations were recognized more than a decade ago~\cite{scheinerman2010rdpg}, yet to the best of our knowledge they have not been satisfactorily addressed in the RDPG literature. 
	Indeed, repeated ASE computation to iteratively impute the unknown entries of $\bbA$ using the inner-product of the embeddings estimated in the previous step lacks convergence guarantees, and multiplies the ASE complexity by the number of iterations~\cite{scheinerman2010rdpg}. \vspace{2pt}
	
	
	\noindent\textbf{Streaming data.} A third scenario that ASE cannot address satisfactorily arises with streaming data from a dynamic network; i.e., when we observe a sequence of graphs over time and would like to track the evolution of the corresponding embeddings, ideally without having to store past observations. Network dynamics may include changes in the edges between a fixed set of nodes (e.g., monitoring a wireless network), the addition of new information (e.g., a user that ranks an item in a recommender system), or the deletion/addition of nodes (e.g., a new user in a social network). Especially for large graphs, re-computing the ASE from scratch each time step is computationally demanding. Given the rotational ambiguity inherent to RDPGs, independently obtaining the ASE after each modification to the graph will likely result in misaligned embeddings that can hinder the assessment of changes. 
	
	\subsection{Contributions and paper outline}
	
	We seek to address these limitations by (i) re-considering the underlying optimization problem of which ASE is a solution (Section \ref{sec:prob_statement}); and (ii) developing iterative embedding algorithms for the refined formulations (Sections \ref{sec:undirected} and \ref{sec:directed}). 
	
	Unlike the ASE recipe of performing a spectral decomposition of $\bbA$, borrowing recent low-rank matrix-factorization advances we advocate solving the non-convex embedding problem using gradient descent (GD)~\cite{Chi2019Review,vu2021icassp}. Explicitly solving the optimization problem facilitates more precise and flexible GRL; e.g., it is straightforward to accommodate unobserved edges by including a mask matrix. Given the iterative nature of GD, warm-restarting the estimates of either new or existing nodes allows to embed multiple (possibly streaming) graphs, while preserving the alignment of consecutive embeddings as a byproduct. 
	Discarding the residuals corresponding to the diagonal of $\bbA$ offers better nodal representations as well as favorable problem structure, which we leverage in Section \ref{sec:coord_desc} to derive block-coordinate descent (BCD) iterations for efficient undirected RDPG inference. 
	
	Applying GD to embed digraph nodes requires special care. As we argue in Section \ref{subsec:interpretability}, inherent ambiguities in the directed RDPG model extend beyond a global rotation, and they may compromise representation quality and the interpretability benefits discussed earlier. We show that an effective way of retaining these desirable features is to impose orthogonality constraints on the matrix factors in the decomposition of $\bbA$ -- a novel GRL formulation for digraphs.  This constraint in turn defines a smooth manifold, over which we optimize using a custom-made feasible method. We stress this is not the well-known Stiefel manifold, where matrices are constrained to be \emph{orthonormal} (and not just orthogonal as here\footnote{We will henceforth use the term \emph{orthonormal matrix} to refer to any matrix $\bbT$ such that $\bbT^\top\bbT=\bbI$ (i.e., the columns of $\bbT$ are orthonormal vectors). The term \emph{orthogonal matrix} will be reserved for those matrices whose columns are mutually orthogonal, but not necessarily of unit norm.}). This is no minor point. Algorithm construction thus requires careful definition of the tangent space, the Riemannian gradient and the retraction~\cite{absil,boumal2023intromanifolds}, all of which we derive in Section \ref{subsec:manifold}. Comprehensive synthetic and real-world (wireless network and United Nations voting) data experiments in Section \ref{sec:applications} demonstrate the interpretability, robustness, and versatility of the novel GRL framework. In the interest of reproducible research, the code and datasets used to generate all figures in this paper is publicly available at \url{https://github.com/marfiori/efficient-ASE}. Concluding remarks are in Section \ref{sec:conclusions}. Non-essential mathematical details {and supplementary experimental results} are deferred to the Appendix. 
	
	All in all, relative to prior art our RDPG embedding framework offers a \emph{better} representation at a \emph{competitive} computational cost, and it is applicable to \emph{more general} settings. This full-blown journal paper extends our preliminary results~\cite{RDPG_EUSIPCO_22,RDPG_Asilomar_22} in multiple significant ways. In addition to a more thorough treatment with extended discussions, unpublished proofs, accompanying software, and expanded numerical studies with new datasets; the BCD algorithm for undirected graphs as well as the treatment of digraphs in Section IV are completely new.
	
	\section{Problem statement and related work}\label{sec:prob_statement}
	
	
	Let us formulate the embedding problem, beginning with undirected graphs. Consider stacking all the nodes' latent position vectors in the matrix $\bbX=[\bbx_1,\ldots,\bbx_N]^\top\in\reals^{N\times d}$. Given an observed graph $\bbA$ and a prescribed embedding dimension $d$ (typically obtained using an elbow rule on $\bbA$'s eigenvalue scree plot~\cite{graspy}), the goal is to estimate $\bbX$. Recalling the definition of the RDPG model, the edge-wise formation probabilities are the entries $P_{ij}=\bbx_i^\top \bbx_j$ of the rank-$d$, PSD matrix $\bbP=\bbX\bbX^\top$. Since we do not allow for self-loops, the diagonal entries in $\bbA$ should be zero and we thus have $\E{\bbA\given \bbX}=\bbM \circ \bbP$, where $\circ$ is the entry-wise or Hadamard product and $\bbM=\mathbf{1}_N\mathbf{1}_N^\top-\bbI_N$ is a mask matrix with ones everywhere except in the diagonal where it is zero. 
	
	In lieu of a maximum likelihood estimator that {is computationally challenging and may not be unique~\cite{xie2023rdpgonestep}}, here we advocate a least-squares (LS) approach~\cite{scheinerman2010rdpg} to obtain
	\begin{gather}\label{eq:ase_mask}
		\hbX\in\argmin_{\bbX\in \reals^{N\times d}}\left\|\bbM \circ (\bbA-\bbX\bbX^\top)\right\|_F^2.
	\end{gather}
	In words, $\hbP=\hbX\hbX^\top$ is the best rank-$d$ PSD approximation to the off-diagonal entries of the adjacency matrix $\bbA$, in the Frobenius-norm sense. The RDPG model is only identifiable up to rotations, and accordingly the solution of \eqref{eq:ase_mask} is not unique. Indeed, for any orthonormal matrix $\bbT\in\reals^{d\times d}$ we have that $\bbX\bbT(\bbX\bbT)^\top=\bbX\bbX^\top=\bbP$.
	
	Entrywise multiplication with $\bbM=\mathbf{1}_N\mathbf{1}_N^\top-\bbI_N$ effectively discards the residuals corresponding to the diagonal entries of $\bbA$. If suitably redefined, the binary mask $\bbM$ may be used for other purposes, such as modeling unknown edges if data are missing. For instance, in a recommender system we typically have the rating of each user over a limited number of items. This (dis)information may be captured in \eqref{eq:ase_mask} by zeroing out the entries of $\bbM$ corresponding to the unknown edges. 
	
	
	\begin{remark}[Adjacency Spectral Embedding]\label{rem:ASE}\normalfont Typically the mask $\bbM$ is ignored (and sometimes non-zero values are iteratively imputed to the diagonal of $\bbA$~\cite{scheinerman2010rdpg}), which results in a closed-form solution for $\hbX$. Indeed, if we let $\bbM=\mathbf{1}_N\mathbf{1}_N^\top$ in \eqref{eq:ase_mask}, we have that $\hbX = \hbV\hbLambda^{1/2}$, where $\bbA=\bbV\bbLambda\bbV^\top$ is the eigendecomposition of $\bbA$, $\hbLambda\in\reals^{d\times d}$ is a diagonal matrix with the $d$ largest-magnitude eigenvalues of $\bbA$, and $\hbV\in\reals^{N\times d}$ are the associated eigenvectors. This workhorse estimator is known as the Adjacency Spectral Embedding~(ASE); see also~\cite{priebe2018survey} for consistency and asymptotic normality results, as well as applications of statistical inference with RDPGs.
	\end{remark}
	
	Given the ASE limitations outlined in Section \ref{subsec:ase_challenges}, we develop efficient gradient-based iterative solvers for the embedding problem \eqref{eq:ase_mask}. Beyond scalability, the algorithmic framework offers a natural pathway to facilitate embedding graph sequences. In the applications we study in Section \ref{sec:applications}, said dynamic network data may be only partially observed, they could be acquired in a streaming fashion, while both the number of nodes and edges are allowed to vary across time. \vspace{2pt}
	
	\noindent \textbf{Embedding digraphs.} Moving on to digraphs~\cite{marques2020digraph}, recall that we now embed each node with two vectors, $\bbx_i^l,\bbx_i^r\in\mathcal{X}\subset\reals^{d}$. Existence of a directed edge from node $i$ to $j$ is modeled as the outcome of a Bernoulli trial with success probability $(\bbx_i^l)^\top\bbx_j^r$~\cite{priebe2018survey}. Again, vertically stacking the embeddings into two matrices $\bbX^l,\bbX^r\in\reals^{N\times d}$, we introduce the probability matrix $\bbP=\bbX^l(\bbX^r)^\top$ such that the expected value of the random adjacency matrix is $\E{\bbA\given \bbX^l,\bbX^r}=\bbM \circ \bbP$.


	If we ignore the mask $\bbM$, the embedding problem boils down to finding the best rank-$d$ approximation to the (possibly asymmetric) adjacency matrix. One such solution may be obtained via the singular value decomposition (SVD) of $\bbA$; i.e.,\ $\bbA=\bbU\bbSigma\bbV^\top$. We thus have that $\hbX^l = \hbU\hbSigma^{1/2}$ and $\hbX^r = \hbV\hbSigma^{1/2}$, with $\hbSigma$ containing only the $d$ largest singular values, and $\hbU$ and $\hbV$ the associated singular vectors. This procedure is also known as ASE. 
	
	As noted in~\cite{marenco2021tsipn}, ambiguities with directed RDPGs can be more problematic than in the undirected case. Now given \emph{any} invertible matrix $\bbT$ (not necessarily orthonormal), the embeddings $\bbY^l=\bbX^l\bbT$ and $\bbY^r=\bbX^r\bbT^{-\top}$ generate the same probability matrix as before; i.e.,\ $\bbY^l(\bbY^r)^\top=\bbX^l\bbT(\bbX^r\bbT^{-\top})^\top=\bbX^l(\bbX^r)^\top=\bbP$. As we show in Section \ref{subsec:interpretability}, constraining matrices $\bbX^l$ and $\bbX^r$ to being orthogonal and having the same column-wise norms\footnote{Let $\bar{\bbx}_i^l, \bar{\bbx}_i^r\in\reals^{N}$ be the $i$-th columns of $\bbX^l$ and $\bbX^r$, respectively. When we say $\bbX^l$ and $\bbX^r$ have the same column-wise norms we mean that $\|\bar{\bbx}_i^l\|_2=\|\bar{\bbx}_i^r\|_2$ holds for all $i=1,\ldots,d$.} effectively limits this ambiguity without compromising the model's expressivity (now an admissible $\bbT$ may only be orthonormal; see Proposition \ref{prop:orthogonal}), all while preserving its interpretability. Given these considerations, our approach to embedding digraphs is to solve the following manifold-constrained optimization problem
	%
	\begin{align}
		\{\hbX^l,\hbX^r\}\in & \argmin_{\{\bbX^l,\bbX^r\}\in \reals^{N\times d}}\left\|\bbM \circ (\bbA-\bbX^l(\bbX^r)^\top)\right\|_F^2\nonumber\\
		\text{s. to } & (\bbX^l)^\top\bbX^l = (\bbX^r)^\top\bbX^r \text{ diagonal}.\label{eq:dase_mask_constrain}
	\end{align}
	In the absence of a mask, a solution of \eqref{eq:dase_mask_constrain} is the legacy ASE. Indeed, $\hbX^l$ and $\hbX^r$ are obtained from orthonormal singular vectors and have the same column-wise norms because both $\hbU$ and $\hbV$ are right-multiplied by $\hbSigma^{1/2}$. To tackle the general case, a novel Riemannian GD  method over the manifold of matrices with orthogonal columns is developed in Section \ref{subsec:manifold}.

	\subsection{Related work}\label{ssec:related_work}
	
	The low-rank matrix factorization problem \eqref{eq:ase_mask} has a long history, with applications to recommender systems (where the objective is to complete a matrix of user-item ratings which is assumed  to have low rank)~\cite{koren2009matrix}; or, in sensor localization from pairwise distances (the so-called Euclidean distance matrix)~\cite{dokmanic2015euclidean}, just to name a couple of examples. Solution methods typically rely on spectral decomposition of the full data matrix (as in ASE), or by considering a convex relaxation via nuclear-norm minimization~\cite{davenport2016overview}. The latter is not best suited for our problem, where we are interested in the actual factors (not in $\bbP$), and their dimensionality could change with time due to e.g., node additions. Alternatively, over the last few years we have witnessed increased interest in non-convex optimization approaches for matrix factorization problems~\cite{Chi2019Review}. Our work may be seen as an effort in this direction. In particular, we bring to bear recent advances in first-order methods for matrix factorization problems and demonstrate impact to GRL (specifically, RDPG inference). The formulation \eqref{eq:dase_mask_constrain} is novel to the best of our knowledge. To solve it we derive GD iterations over the manifold of orthogonal matrices, which is different from the Stiefel manifold and thus requires careful treatment given the unique geometric properties of our problem. 
	
	The scalability of ASE, or any other spectral embedding method for that matter, has long been considered an issue~\cite{brand2006fast}. This challenge is compounded when multiple graphs are to be embedded, especially in \emph{batch} settings where all graphs in the sequence are stored in memory~\cite{gallagher2021spectral}. Existing approaches seeking aligned embeddings rely on the spectral decomposition of some matrix whose dimension(s) grow linearly with the number of graphs~\cite{levin2017omnibus,jones2020multilayer,gallagher2021spectral}. In addition to increasing the computation cost of ASE, these methods are not applicable in streaming scenarios, where a possibly infinite sequence of graphs $\{\bbA_t\}$ is observed and we want to recursively update the embeddings `on-the-fly' as new graphs are acquired.  
	
	There is an extensive collection of numerical linear algebra approaches to recursively update an eigendecomposition or SVD when the (adjacency) matrix is perturbed; e.g.,~\cite{brand2006fast}. However, these do not offer major computational savings except for specific types of changes (e.g., rank-$1$ updates), and they may be prone to error accumulation as $t$ increases~\cite{zhang2018timers}. Moreover, they can yield misaligned embeddings due to the rotational ambiguity of RDPGs. The sketching literature offers highly-scalable randomized algorithms~\cite{tropp2011sketching}. Other than to initialize our iterative methods we do not consider those here, because we are interested in exact solutions to \eqref{eq:ase_mask} and \eqref{eq:dase_mask_constrain}.
	
	In dynamic environments, not only does $\bbA_t$ change over time, but new nodes may be added to the graph and current ones removed. Embedding previously unseen nodes corresponds to an inductive learning setting, generally regarded as beyond the capabilities of shallow embeddings as the one we are discussing here~\cite[Ch. 3.4]{hamilton2020representation},~\cite{chami2022taxonomy}. Previous efforts in this direction (that avoid re-computing eigendecompositions from scratch) either assume that the connections between the existing nodes, or, their current embeddings, do not change~\cite{pana2023icassp,oos2018levin}. In the latter case, a projection scheme onto the space of current embeddings produces an asymptotically $(N\to\infty)$ consistent ASE for the new node~\cite{oos2018levin}. However, even if latent positions were time invariant, the estimation error of current nodes' embeddings propagates to the new estimates. We will use the projection-based estimate in~\cite{oos2018levin} as initialization for new nodes in our GD algorithms, demonstrating benefits in accuracy and stability especially as several nodes are added, while at the same time refining previous nodes' embeddings.

	\section{Embedding Algorithms for Undirected Graphs}\label{sec:undirected}
	
	We start with the embedding problem for undirected graphs. Recognizing limitations of state-of-the-art ASE implementations, here we first review a GD algorithm with well-documented merits for symmetric matrix completion, yet so far unexplored in RDPG inference. GD offers flexible computation of embeddings and a pathway towards tracking nodal representations in a streaming graph setting. We then show that the particular structure of the problem lends itself naturally to more efficient BCD iterations, and discuss the relative merits of the different approaches in terms of convergence properties, complexity, and empirical execution time.
	
	\subsection{Back to basics: Estimation via gradient descent}\label{sec:grad_desc}
	
	Recall the embedding problem for undirected graphs in \eqref{eq:ase_mask}, and denote by $f:\reals^{N\times d} \to \reals$ its smooth objective function $f(\bbX) = \|\bbM \circ (\bbA-\bbX\bbX^\top)\|_F^2$. Although the problem is not convex with respect to $\bbX$,  it is convex with respect to $\bbP=\bbX\bbX^\top$. 
	In the broad context of matrix factorization problems where the objective function depends on the product $\bbX\bbX^\top$, the GD approach is often referred to as \textit{factored GD}~\cite{bhojanapalli2016}.
	
	The workhorse GD algorithm generates embedding updates 
	\begin{equation}\label{eq:gd}
		\bbX_{k+1} = \bbX_k - \alpha \nabla f(\bbX_k),\quad k=0,1,2,\ldots
	\end{equation}
	where $\alpha >0$ is the stepsize, and the gradient of $f$ is $\nabla f(\bbX) = 4\left[\bbM\circ (\bbX\bbX^\top-\bbA)\right ] \bbX$.  Recall that $\bbA$ and $\bbM$ are known symmetric matrices that specify the problem instance.
	
	There have been several noteworthy advances in the study of GD's convergence (including rates) for this non-convex setting, as well as accelerated variants~\cite{chen2015,bhojanapalli2016,sun2016,zhou2020,Chi2019Review,vu2021icassp}.
	For the RDPG embedding problem dealt with here, it follows that if the initial condition $\bbX_0$ is close to the solution of \eqref{eq:ase_mask}, the GD iteration \eqref{eq:gd} converges linearly to $\hbX${; see \cite[Corollary 7]{bhojanapalli2016}, \cite[Theorem 1]{vu2021icassp}, as well as \cite[Lemma 4]{Chi2019Review} and references therein for a similar version of this proposition.}
	\begin{myproposition}
		\label{prop:convergence}
		Let $\hbX$ be a solution of \eqref{eq:ase_mask}. Then there exist $\delta >0$ and $0<\kappa<1$ such that, if $d(\bbX_0,\hbX)\leq \delta$, we have
		\begin{gather}
			d(\bbX_k,\hbX) \leq \delta \kappa^k  , \quad \forall\: k >0
		\end{gather}
		where $\{\bbX_k\}$ are GD iterates \eqref{eq:gd} with appropriate constant stepsize, and $d(\bbX,\hbX) := \min_{\bbW } \|\bbX\bbW - \hbX\|_F^2$  s. to 
		$\bbW^\top\bbW=\bbW\bbW^\top=\bbI_d$ is a matrix distance accounting for the rotational invariance.
	\end{myproposition}
	Although there are specific $\bbX_0$ which correspond to sub-optimal stationary points, in our experience GD converges to the global optimum when initialized randomly. {For further details on the initialization of factored GD, strong convexity assumptions, and the choice of the stepsize $\alpha$, see e.g., \cite[Section 5]{bhojanapalli2016}.}
	
	\begin{remark}[Warm restarts to embed multiple graphs]\label{rem:warm_restarts}\normalfont On top of being flexible to handle missing data encoded in $\bbM$, this approach also allows to track the latent positions of a graph sequence $\{\bbA_t\}$ using warm restarts. That is, after computing the embeddings $\bbX_t$ of the graph with adjacency matrix $\bbA_t$, we initialize the GD iterations \eqref{eq:gd} with $\bbX_t$ to compute $\bbX_{t+1}$ corresponding to $\bbA_{t+1}$. This way, unless the graphs at times $t$ and $t+1$ are uncorrelated, the embedding of each node will transition smoothly to its new position (up to noise). {Moreover, if the embeddings of the graph at time $t+1$ are sufficiently close to the embeddings at time $t$, say for slowly time-varying graphs where $d(\bbX_t,\bbX_{t+1})\leq \delta$, then the GD iterates for computing $\bbX_{t+1}$ also have the same convergence guarantees and rates ($\delta\kappa^k$), by virtue of Proposition \ref{prop:convergence}. This was also observed empirically, indeed } experiments demonstrating this longitudinal stability property~\cite{gallagher2021spectral} are presented in Sections \ref{sec:warm} and \ref{sec:tracking}.
	\end{remark}

	\subsection{Block coordinate descent}\label{sec:coord_desc}
	
	Here we develop a BCD method for solving \eqref{eq:ase_mask}, which turns out to be quite efficient. The algorithm updates one row of $\bbX$ at a time in a cyclic fashion, by minimizing the objective function $f$ with respect to the corresponding row (treating all other entries of $\bbX$ as constants, evaluated at their most recent updates). In general, this row-wise sub-problem may not admit a simple solution; however, we show that due to the structure of the mask matrix $\bbM$ the updates are given in closed form.
	
	Let $f(\bbX) = \|\bbM \circ (\bbA-\bbX\bbX^\top)\|_F^2$ and recall $\bbx_i^\top$ is the $i$-th row of $\bbX$. The gradient $\nabla_i f$ of $f$ with respect to $\bbx_i$  is 
	\begin{align}
		\nabla_if(\bbX) &= \left(-(\bbM \circ \bbA)_i \bbX + ((\bbM\circ \bbX\bbX^\top)\bbX)_i\right)^\top\nonumber \\
		&= -\bbX^\top(\bbM \circ \bbA)_i^\top  + \bbX^\top(\bbM\circ \bbX\bbX^\top)_i^\top,\label{eq:grad_row_i}
	\end{align}
	where $(\cdot)_i$ stands for the $i$-th row of the matrix argument.
	
	Note that since the graph has no self-loops (i.e., the diagonal entries of $\bbA$ are zero), then the entry-wise product of $\bbA$ with $\bbM$ is vacuous over the diagonal. Also because $A_{ii}=0$, the term $\bbX^\top(\bbM \circ \bbA)_i^\top  = \bbX^\top(\bbA)_i^\top$ in \eqref{eq:grad_row_i} does not depend on $\bbx_i$. More importantly, $\bbX\bbX^\top$ clearly depends on $\bbx_i$, and this would challenge solving $\nabla_if(\bbX) = \bb0_d$ to obtain a minimizer [due to the trilinear form of the second term in \eqref{eq:grad_row_i}]. However, a close re-examination of $\bbX^\top(\bbM\circ \bbX\bbX^\top)_i^\top$ suggests this purported challenge can be overcome. First, observe that 
	$$(\bbM\circ \bbX\bbX^\top)_i =\bbx_i^\top\bbX^\top - \bbr_i^\top,$$
	where $\bbr_i \in \reals^N$ is a column vector with zeros everywhere except in entry $i$, where it takes the value $\bbx_i^\top\bbx_i$. Hence,
	%
	%
	\begin{equation*}
		\bbX^\top(\bbM\circ \bbX\bbX^\top)_i^\top   
		= \bbX^\top(\bbX\bbx_i-\bbr_i)
		= 
		(\bbX^\top\bbX - \bbx_i\bbx_i^\top)\bbx_i.
	\end{equation*}
	All in all, we have a simplified expression for the gradient
	\begin{equation}
		\nabla_if(\bbX) = -\bbX^\top (\bbA_i)^\top + (\bbX^\top\bbX - \bbx_i\bbx_i^\top)\bbx_i.\label{eq:grad_row_simplified}
	\end{equation}
	Now, define $\bbR = \bbX^\top\bbX - \bbx_i\bbx_i^\top$ and notice this matrix does not depend on $\bbx_i$. Therefore, from \eqref{eq:grad_row_simplified} it follows that the equation $\nabla_if(\bbx_i)=\bb0_d$ is \emph{linear} in $\bbx_i$, namely $\bbR\bbx_i = \bbX^\top(\bbA_i)^\top$. The pseudo-code of the algorithm is tabulated under Algorithm \ref{algo:coord_descent}.
	
	The $d\times d$ matrix $\bbR$ is invertible provided that $\bbX$ has rank $d$. This also implies that the row-wise sub-problem has a unique minimizer, which is key to establish convergence of BCD to a stationary point~\cite[Prop. 2.7.1]{bertsekas99}. It is worth reiterating that this favorable linear structure is lost in the absence of a mask matrix $\bbM$ (cf. ASE in Remark \ref{rem:ASE}).  Since in RDPG embeddings we  typically have $d \ll N$, solving multiple $d\times d$ linear systems is affordable; especially when compared to matrix-vector operations of order $\Theta(N^\gamma)$, $\gamma>1$, like in GD.
	
	
	\begin{algorithm}[t]
		\caption{Block coordinate descent (BCD)}
		\label{algo:coord_descent}
		\algsetup{linenosize=\normalsize}
		\begin{algorithmic}[1]
			\REQUIRE Initial $\bbX \gets \bbX_0$
			\STATE Compute $\bbR = \bbX^\top\bbX$
			\REPEAT
			\FOR{$i=1:N$}
			\STATE $\bbR \gets \bbR - \bbx_i \bbx_i^\top$
			\STATE $\bbb \gets \bbX^\top(\bbA_i)^\top$
			\STATE $\bbx_i \gets $ solution of $\bbR\bbx=\bbb$
			\STATE $\bbR \gets \bbR + \bbx_i \bbx_i^\top$
			\ENDFOR
			\UNTIL{convergence}
			\RETURN $\bbX$.
		\end{algorithmic}
	\end{algorithm}
	
	\subsection{Complexity and execution time analyses}\label{sec:complexity}
	
	We compare four computational methods to obtain RDPG embeddings of undirected graphs: the ASE based on (i) full eigendecomposition, and (ii) truncated SVD as implemented in \texttt{Graspologic}~\cite{graspy}; 
	(iii) GD initialized with the randomized-SVD (RSVD)~\cite{tropp2011sketching} (we  account for the RSVD in the execution time); and (iv) randomly initialized BCD as in Algorithm \ref{algo:coord_descent}.
	
	The full eigendecomposition of $\bbA$ has worst-case $\Theta(N^3)$ complexity, while for sparse graphs the $d$-dominant components can be obtained with $\Theta(Nd)$ per-iteration cost. For GD, the per-iteration computational cost incurred to evaluate $\nabla f(\bbX)$ is dominated by the matrix multiplications, which is $\Theta(N^2d)$ for a na\"ive implementation. The number of iterations depends on $\bbX_0$, but even with a favorable initialization the runtime is still higher than the $\Theta(Nd)$ of truncated SVD-based ASE. A refined convergence-rate analysis of GD for the symmetric matrix completion problem is presented in~\cite{vu2021icassp}. 
	
	Although in general it is tricky to compare the complexity of GD against BCD approaches, we can evaluate the per-iteration computational cost of both methods (for BCD this means a whole cycle over the rows of $\bbX$ is completed). In both cases, each entry of the matrix $\bbX$ is updated exactly once. Each cycle consist of $N$ instances of $d\times d$ linear systems, so this is $\Theta(Nd^3)$ in the worst case. In addition, in our experience Algorithm \ref{algo:coord_descent} converges in fewer iterations than the GD method.
	
\begin{figure}[t!]
		\centering
		\includegraphics[width=0.5\textwidth]{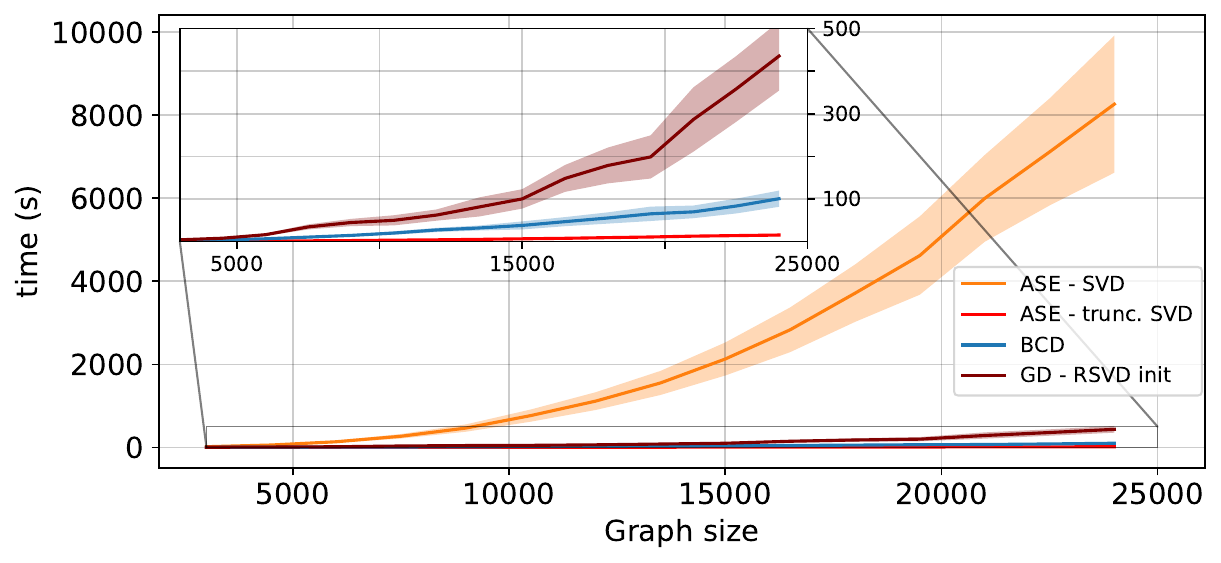}
		\caption{Execution time for embedding SBM graphs with up to $N=24000$ nodes. As $N$ grows, BCD exhibits competitive scaling to the state-of-the-art ASE algorithm implemented in the \texttt{Graspologic} package.}
		\label{fig:complexity}
	\end{figure}

	In Fig. \ref{fig:complexity} we compare the execution times of methods (i)-(iv) as a function of $N${, all the way to $N=24000$}. For ASE, we use the \texttt{SciPY} optimized implementation of the eigendecomposition in Python, as in state-of-the-art RDPG inference packages such as \texttt{Graspologic}~\cite{graspy}. Our GD and BCD implementations are in pure Python, not optimized for performance.  
	For each $N$, we sampled several 2-block SBM graphs, with connection probabilities of $p=0.5$ (within block) and $q=0.2$ (across blocks). Community sizes are $N/3$ and $2N/3$. We let $d=2$ in all cases. Results are averaged over $10$ Monte Carlo replicates, and corresponding standard deviations are depicted in Fig. \ref{fig:complexity}. In all cases, the methods converge to a solution of \eqref{eq:ase_mask}. The obtained cost function is very similar for each run, with slightly lower values for the GD and BCD methods because they are solving the problem with the zero-diagonal restriction. As expected, BCD exhibits competitive scaling with the truncated SVD-based ASE, and can embed graphs with $N=20000$ nodes in just over a minute. 

	{A moderately large graph, such as the one with $N=24000$, is ideal to assess the effect of $d$ on the computation time. Graphs of this scale are expected to have several communities, and thus values larger than $d=2$ (as before) will likely be required. We thus explore this scenario in more detail, embedding a $d$-block SBM graph using $d$ dimensions, and measure the execution time of BCD (Algorithm \ref{algo:coord_descent}) and the truncated SVD methods as $d$ increases. Results are reported in Table \ref{tab:tiempo_d_grande}. Interestingly, BCD yields faster results than truncated SVD when $d\geq 50$, gracefully scaling with the embedding dimension. As we mentioned before, in our experience BCD converges in very few iterations and offers competitive computation performance.}

	\begin{table}
		\centering
		{
			\caption{Execution time (in seconds) as a function of the embedding dimension $d$, for $d$-block SBM graphs with $N=24000$ nodes.}}
		\label{tab:tiempo_d_grande}
		{\begin{tabular}{|c||c|c|c|}
				\hline
				& $d=10$ & $d=50$ & $d=100$\\ \hline\hline
				BCD - Algorithm \ref{algo:coord_descent} & $67 \pm 4$ & $98 \pm 9$ & $154 \pm 32$\\ \hline
				ASE - Truncated SVD & $14 \pm 1$ & $247 \pm 21$ & $466 \pm 47$ \\
				\hline
		\end{tabular}}
	\end{table}

	\section{Embedding Algorithms for Digraphs}\label{sec:directed}
	
	Shifting gears to embedding nodes in digraphs, we start with a close examination of the ambiguities inherent to the directed RDPG model and justify the need for orthogonality constraints on the factors' columns. To compute the desired nodal representations, we then develop a first-order feasible optimization method in the manifold of matrices with orthogonal columns.
	
	\subsection{On the interpretability of the directed RDPG}\label{subsec:interpretability}
	Recall that if $\{\hbX^l,\hbX^r\}$ is a minimizer of $f(\bbX^l,\bbX^r)=\|\bbM\circ(\bbA-\bbX^l(\bbX^r)^\top)\|_F^2$, then so is $\{\hbX^l\bbT,\hbX^r\bbT^{-\top}\}$ for any invertible matrix $\bbT$. Let us now discuss why $\bbX^l$ and $\bbX^r$ should be constrained to be orthogonal and have the same column-wise norms. In other words, why do we need the constraints in \eqref{eq:dase_mask_constrain} to obtain useful embeddings when the graph is directed.
	
	To gain some insights, suppose we ignore these constraints altogether and use GD to minimize $f(\bbX^l,\bbX^r)$. Similar to \eqref{eq:gd}, at iteration $k+1$ we update  $\{\bbX^l_{k+1},\bbX^r_{k+1}\}$ as follows
	\begin{gather}
		\bbX^l_{k+1} = \bbX^l_k - \alpha \nabla_{\bbX^l} f(\bbX^l_k,\bbX^r_k),\label{eq:dgdl}\\
		\bbX^r_{k+1} = \bbX^r_k - \alpha \nabla_{\bbX^r} f(\bbX^l_k,\bbX^r_k),\label{eq:dgdr}
	\end{gather}
	where $\nabla f_{\bbX^l}(\bbX^l,\bbX^r) = 4\left[\bbM\circ (\bbX^l(\bbX^r)^\top-\bbA)\right] \bbX^l$ and a similar expression holds for $\nabla f_{\bbX^r}(\bbX^l,\bbX^r)$. 
	
	The ASE offers an alternative baseline, which requires to discard the mask $\bbM$. ASE estimates $\{\hbX^l,\hbX^r\}$ have orthogonal columns because they are derived from the SVD of $\bbA$. Same index columns in $\hbX^l$ and $\hbX^r$ have the same norm as well, since the orthonormal matrices $\hbU$ and $\hbV$ are \emph{both} right-multiplied by $\hbSigma^{1/2}$. 
	However, if we minimize $f(\bbX^l,\bbX^r)$ iteratively as in \eqref{eq:dgdl}-\eqref{eq:dgdr} to accommodate missing and streaming data, we may loose column-wise orthogonality with detrimental consequences we illustrate in the following example.
	
	\begin{myexample}[Bipartisan senate]\label{Ex:senate}\normalfont Consider a synthetic political dataset of votes cast by senators in support of laws, over a certain period of time. This may be regarded as a bipartite
		digraph where nodes correspond to senators and laws, and the fact that senator $i$ has voted affirmatively for law $j$ is indicated by the existence of edge $(i,j)$. 
		We examine a polarized scenario, where two political parties put forth laws for voting. Affirmative votes are more likely for senators from the party that introduced the law, and less likely for senators from the opposing party. There are also a few bipartisan laws, for which affirmative votes tend to be more balanced across parties. We simulated such a graph with 50 senators of each party, {where Party 1 proposed 50 laws and Party 2 proposed 200 laws, and there were 40 additional bipartisan laws under consideration (i.e., $N=390$ in total). 
			Furthermore, the inter-community probability matrix is
			\begin{gather*}
				\bbPi = \begin{psmallmatrix}
					0 & 0 & 0.9 & 0.01 & 0.2 \\
					0 & 0 & 0.1 & 0.8 & 0.3 \\
					0 & 0 & 0 & 0 & 0 \\
					0 & 0 & 0 & 0 & 0 \\
					0 & 0 & 0 & 0 & 0 \\
				\end{psmallmatrix},
			\end{gather*}
			where communities are ordered as Party 1 senator, Party 2 senator, Party 1 law, Party 2 law, and finally bipartisan law.
			In other words, senators of Party 1 are very supportive of their own laws and unlikely to vote for those introduced in the other side of aisle, whereas Party 2 senators are more moderate.}
		
		
		We compare the embeddings estimated through ASE and by GD [i.e., iterating \eqref{eq:dgdl} and \eqref{eq:dgdr} until convergence]. {Recall that interpretation of these results should rely on the geometry induced by the RDPG model. Similarity among nodes is captured by their colinearity in latent space, not by their Euclidean distance being small (as it is the case with e.g., Laplacian eigenmaps~\cite{laplacian_eigenmaps}). Accordingly, in this particular example we expect that the embeddings of Party 1 senators and laws will be almost perfectly aligned, while slightly less so for Party 2.} ASE results using $d=2$ are shown in Figure~\ref{fig:embeddings_bipartidismo_no_ortogonal} (left). As expected, the outward embeddings for laws and inward embeddings for senators are  zero (since the former do not vote and the latter are not voted). Furthermore, the outward embeddings corresponding to senators of each party are close and roughly orthogonal to senators of the other party, reflecting the polarized landscape. Finally, the inward embeddings of laws submitted by each party are aligned with the corresponding cluster of senators, whereas bipartisan laws lie somewhere between both groups. {The difference in magnitude between embeddings of senators and laws is due to the different number of such nodes in the graph, and the column-wise norm constraint imposed to the embeddings.}
		
		
		\begin{figure}
			\centering
			\includegraphics[width=\linewidth]{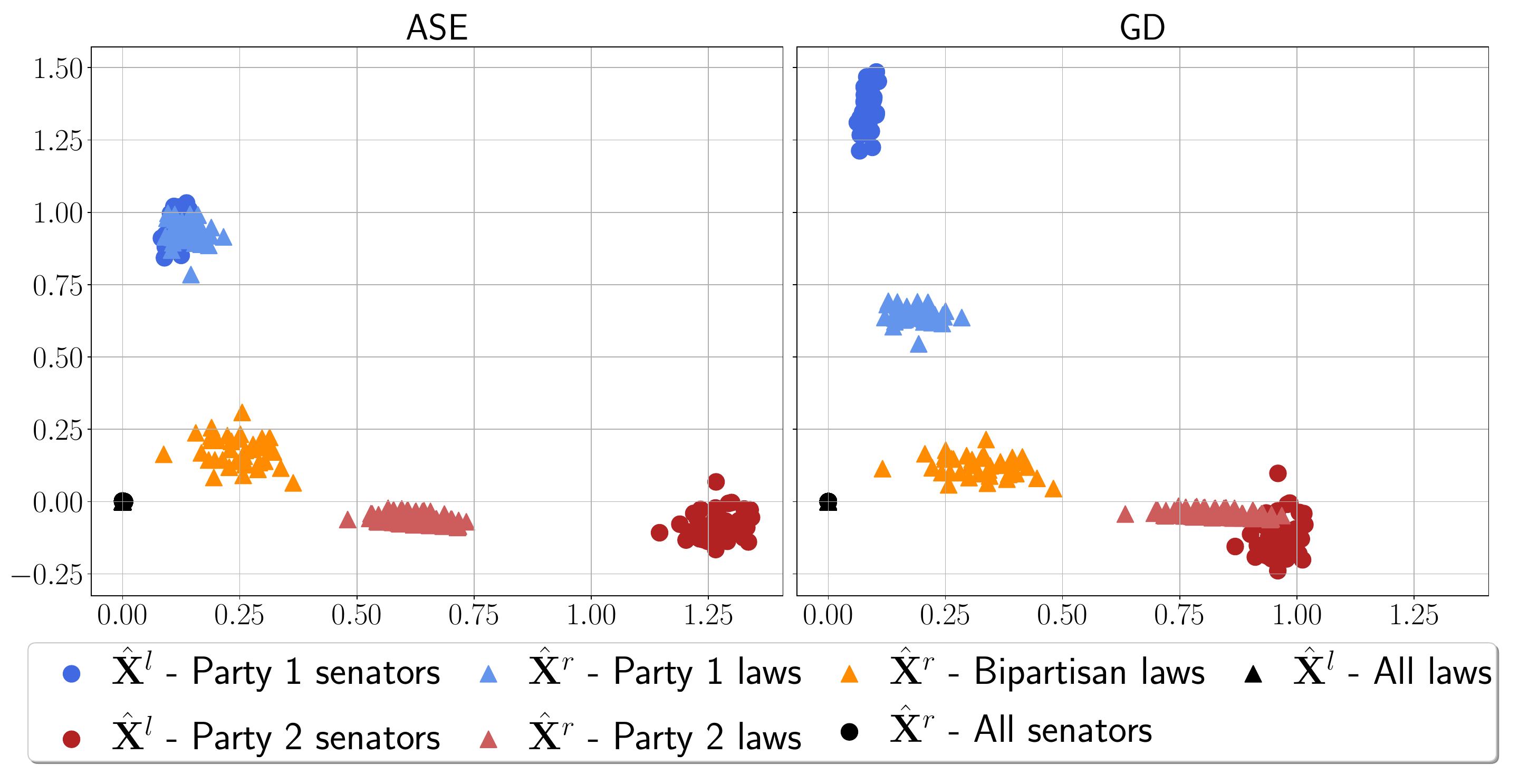}
			\caption{
				Bipartisan senate example. ASE (left) and GD (right). Since ASE implicitly imposes equally normed orthogonal columns (as it is derived from an SVD), it produces interpretable embeddings. On the other hand, GD may result in laws and parties that are not aligned, and thus loses interpretability if no further constrains are imposed in the formulation.}
			\label{fig:embeddings_bipartidismo_no_ortogonal}
		\end{figure}
		
		On the other hand, inspection of Fig. \ref{fig:embeddings_bipartidismo_no_ortogonal} (right) reveals that GD converges to a solution where laws are not aligned with the corresponding party senators. Accordingly, the affinity of parties to their laws is less evident than before. In fact, it appears as if Party 1 is not as supportive of its laws as in the ASE-based visualization {which, as we discussed before, is the opposite of what we expected}. While the input graph is the same for both methods and the total cost $f(\hbX^l,\hbX^r)$ is smaller for the GD method, interpretability is hindered because of the larger ambiguity set in the absence of additional constraints.
	\end{myexample}
	
	\begin{myexample}[Digraph with symmetric expectation]\label{Ex:directed_sym_P} \normalfont To further justify why orthogonality constraints are essential, consider a digraph sampled from a symmetric $\bbP$ (i.e., the probability of both edge directions is the same, but each arc is independently sampled). It would be desirable that in this case the model enforced $\bbX^l=\bbX^r$, since the outgoing and incoming behaviour of the nodes is the same. The general directed model should recover the subsumed undirected one and, naturally, the same should hold for the embedding method. 
		
		However, these desiderata are not necessarily met. As stated earlier, given an invertible matrix $\bbT$, the embeddings $\bbY^l=\bbX^l\bbT$ and $\bbY^r=\bbX^r\bbT^{-\top}$ yield the same probability matrix $\bbP=\bbX^l(\bbX^r)^\top$. This impies that unless $\bbT=\bbT^{-\top}$ (meaning $\bbT$ is an orthonormal matrix  corresponding to a rotation), we could apply \emph{different} transformations to the inward and outward embeddings and still obtain the same RDPG. 
	\end{myexample}
	
	Given these observations, consider a directed RDPG model where the embedding matrices $\bbX^l$ and $\bbX^r$ are constrained to be orthogonal and of the same column-wise norm. The following result asserts that this suffices to ensure an admissible $\bbT$ is orthonormal, hence reducing the model's ambiguity to a global rotation -- just like in the undirected case. 
	\begin{myproposition}
		\label{prop:orthogonal}
		Let $\bbP=\bbX^l(\bbX^r)^\top$ be the probability matrix of a directed RDPG model, where $\{\bbX^l,\bbX^r\}$ are $N\times d$ matrices with rank $d$ such that $(\bbX^l)^\top\bbX^l=(\bbX^r)^\top\bbX^r=\bbD_X$ is diagonal. Let $\bbT\in\reals^{d\times d}$ be an invertible matrix such that $\bbY^l=\bbX^l\bbT$ and $\bbY^r=\bbX^r\bbT^{-\top}$ 
		are also orthogonal with the same column-wise norms; i.e.\  $(\bbY^l)^\top\bbY^l = (\bbY^r)^\top\bbY^r=\bbD_Y$ is diagonal. 
		Then, $\bbT$ is an orthonormal matrix.
	\end{myproposition}
	\begin{myproof}
		Combining $\bbY^r=\bbX^r\bbT^{-\top}$ with $(\bbX^r)^\top\bbX^r=\bbD_X$ and $(\bbY^r)^\top\bbY^r=\bbD_Y$ we find that $\bbD_X = \bbT \bbD_Y \bbT^\top$. Proceeding analogously with $\bbY^l$ we further obtain that $\bbD_X = \bbT^{-\top} \bbD_Y \bbT^{-1}$. Multiplying both identities results in $\bbD_X^2 = \bbT\bbD_Y^2\bbT^{-1}$. Thus, the columns of $\bbT$ are linearly independent eigenvectors of a diagonal matrix. Furthermore, given $\bbD_X = \bbT \bbD_Y \bbT^\top$ it follows that the above eigendecomposition is necessarily one with orthonormal eigenvectors. 
	\end{myproof}

	The constraints in \eqref{eq:dase_mask_constrain} do not limit the expressiveness of the model, since they are compatible with those ASE implicitly imposes. Next, we develop a feasible first-order method that enforces the orthogonality constraint at all iterations. After convergence it is straightforward to equalize the resulting column-wise norms so that they are the same for both $\hbX^l$ and $\hbX^r$, without affecting the generated $\bbP$; see Remark \ref{rem:scaling_cols}.  
	
	\subsection{Optimizing on a manifold}\label{subsec:manifold}
	We have concluded that for the sake of interpretability and quality of the representation, it is prudent to impose the matrices $\bbX^l$ and $\bbX^r$ have orthogonal columns. One classical way to tackle this is by adding these constraints to the optimization problem as in \eqref{eq:dase_mask_constrain}, and solve it via Lagrangian-based methods.
	For some constraints with geometric properties, a more suitable and timely approach is to pose the optimization problem on a smooth manifold. One can then resort to feasible methods that search exclusively over the manifold, i.e., the constraints are satisfied from the start and throughout the entire iterative minimization process~\cite{absil,boumal2023intromanifolds}. This way, we can think of the optimization as being \emph{unconstrained} because the manifold is all there is. In the sequel we explore this last idea.
	
	Interestingly, the space of matrices having orthogonal columns does not form any known and well-studied manifold. Yet, we show the required geometric structure is present in our problem and thus we have to define several objects as well as compute various operators to facilitate optimization~\cite{absil,boumal2023intromanifolds}. The conceptual roadmap is as follows. Recall that a smooth manifold $\ccalM$ can be locally approximated by a linear space, the so-called \emph{tangent space}. If we consider the objective function $f:\ccalM\mapsto \reals$ defined from the (Riemannian) manifold to $\reals$, then the Riemannian gradient of the function is an element of the tangent space. This \emph{Riemaninann gradient}, which will be denoted as $\grad f$, can be computed as the projection of the Euclidean gradient $\nabla f$ to the tangent space. Having computed the gradient, a classical descent method consists of taking a certain step in the opposite direction. However, this step likely results in a point outside of the manifold, so we have to project it back to $\ccalM$. This projection might be computationally intensive, so the \emph{retraction} alternative is used instead. 
	
	Next, we define more precisely our manifold, and derive the tangent space, the projection and finally the retraction. 
	The manifold that resembles the most to ours is the so-called Stiefel manifold, which consist of matrices with orthogonal and unit-norm (i.e., orthonormal) columns
	\begin{equation}
		St(d,N) := \{\bbX \in \reals^{N\times d}: \bbX^\top \bbX = \bbI_d\}.
	\end{equation}
	But here we do not require unit-norm columns. 
	Thus, let $\reals^N_*=\reals^N \setminus \{\mathbf{0}_N\}$ be the set of $N$ dimensional vectors without the null vector, and let $\reals^{N\times d}_*$ be the product of $d$ copies of $\reals^{N}_*$. This open set is the set of $N \times d$ matrices without null columns.
	We are interested in matrices with orthogonal columns, namely
	\begin{align}
		\mathcal{M}(d,N) := &\:\{\bbX \in \reals^{N\times d}_*: \bbX^\top \bbX \text{ is diagonal}\}\label{eq:fiori_manifold}\\ = &\:\{\bbX \in \reals^{N\times d}_*: \bbM\circ (\bbX^\top \bbX)=\bb0_{d\times d}\},\nonumber
	\end{align}
	where 
	once more $\bbM=\mathbf{1}_d\mathbf{1}_d^\top-\bbI_d$ is a particular mask matrix, with zeros in the diagonal and ones everywhere else. 
	
	
	The following proposition establishes that $\mathcal{M}$ is actually a manifold (for notational convenience, we henceforth use $\mathcal{M}$ instead of $\mathcal{M}(d,N)$ since both $d$ and $N$ are fixed throughout). Moreover, $\mathcal{M}$ is a Riemannian manifold since  $\mathcal{M}\subset \reals^{N\times d}$ is a vector space equipped with the usual trace inner product.
	The proofs of subsequent results can be found in the Appendix. 
	
	\begin{myproposition}\label{stiefel_tilde_manifold}
		The set $\mathcal{M}$ in \eqref{eq:fiori_manifold} is a differential manifold and its dimension is $Nd-d(d-1)/2$. Furthermore, the tangent space at $\bbX \in \mathcal{M}$ is given by 
		\begin{equation*}
			\mathrm{T}_{\bbX}\mathcal{M}  = \{\bbzeta  \in \reals^{N\times d}: \bbM \circ \left( \bbzeta^\top \bbX + \bbX^\top\bbzeta\right)=\mathbf{0}_{N\times d}  \}.
		\end{equation*}
	\end{myproposition}
	
	To perform a manifold GD step, one needs to compute the Riemmanian gradient of the function defined in $\ccalM$. We obtain $\grad f$ as the projection of the Euclidean gradient [cf. \eqref{eq:dgdl}-\eqref{eq:dgdr}] onto the tangent space. 
	A natural way to compute said projection 
	is to first characterize and compute the projection to the normal space.
	Given $\bbX \in \mathcal{M}$, the normal space at $\bbX$ is  $T_{\bbX}\mathcal{M}^\perp = \{\bbN \in \reals^{N\times d}: \langle \bbN,\bbzeta \rangle = \tr(\bbN^\top \bbzeta) =0, \forall \,\bbzeta \in T_{\bbX}\mathcal{M}\}.$ 
	A useful alternative characterization is given next.
	\begin{mylemma}\label{lemma:charact_normal}
		The normal space at $\bbX$ is
		\begin{equation*}
			T_{\bbX}\mathcal{M}^\perp =\{\bbX\bbLambda \in \reals^{N\times d}: \bbLambda \in \mathcal{S}_d\},
		\end{equation*}
		where $\mathcal{S}_d = \{\bbX \in \reals^{d\times d}: \bbX=\bbX^\top, \textrm{diag}(\bbX)=\mathbf{0}_{d\times d}\}$ is the set of $d\times d$ symmetric matrices with null diagonal.
	\end{mylemma}
	Computing the projection to the normal space requires some work {due to the null diagonal constraint in $\mathcal{S}_d$, which is not present in the normal space to $St(d,N)$~\cite[p. 161]{boumal2023intromanifolds}.} The result is given in the next lemma. 
	\begin{mylemma}\label{lemma:projec_normal_space}
		Let $\bbX \in \mathcal{M}$ and let $\pi_\bbX^\perp:\reals^{N\times d}\mapsto T_{\bbX}\mathcal{M}^\perp$ be the projection to the normal space. Then
		\begin{equation}\label{eq:projec_normal}
			\pi_\bbX^\perp(\bbZ) = \bbX s(2\bbD\bbL),
		\end{equation}
		%
		where $s:\reals^{d\times d} \mapsto \mathcal{S}_d$ is a symmetrizing function $s(\bbZ) = \frac{\bbZ + \bbZ^\top}{2} - \textrm{diag}(\bbZ)$,
		$\bbD = \left(\bbX^\top\bbX\right)^{1/2}$ and $\bbL = \left(\bbD^{-1}\bbX^\top \bbZ\right)\circ \bbF$, where $\bbE = \mathbf{1}_d\mathbf{1}_d^\top \bbD^2 + \bbD^2\mathbf{1}_d\mathbf{1}_d^\top$ and $\bbF$ has entries $F_{ij}=E_{ij}^{-1}$. 
	\end{mylemma}
	
	Note that \eqref{eq:projec_normal} is of the form $\bbX\bbLambda$, with $\bbLambda \in \mathcal{S}_d$. It thus belongs to the normal space by virtue of the characterization in Lemma \ref{lemma:charact_normal}. The calculations to show it is indeed the projection are detailed in the Appendix, and boil down to proving that $\bbZ - \pi_\bbX^\perp(\bbZ)$ lives in the tangent space. Specifically, to establish \eqref{eq:projec_normal} we take $\bbX\bbLambda$ and derive conditions that $\bbLambda$ had to verify when imposing that $\bbZ - \bbX\bbLambda\in \mathrm{T}_{\bbX}\mathcal{M}$. After some derivations, we find $\bbLambda=s(2\bbD\bbL)$, with the auxiliary matrices $\bbD, \bbE, \bbF$ and $\bbL$ defined in Lemma \ref{lemma:projec_normal_space}. Finally, the desired projection to the tangent space is given as 
	follows.
	\begin{myproposition}\label{prop:proj_tangent}
		Let $\bbX\in \mathcal{M}$. The projection to the tangent space $\pi_\bbX:\reals^{N\times d}\mapsto T_{\bbX}\mathcal{M}$ can be computed as:
		\begin{equation*}
			\pi_\bbX(\bbZ) =\bbZ -\pi_\bbX^\perp(\bbZ) = \bbZ - \bbX s(2\bbD\bbL).
		\end{equation*}
	\end{myproposition}

	When we take a small step in the opposite direction of $\grad f$, in general we fall outside $\ccalM$ and we have to project back to it. We need a projection from the tangent bundle to the manifold, or a retraction, which is more efficient in general.
	
	Given a full rank matrix $\bbZ \in \reals^{N\times d}$, consider its decomposition $\bbZ = \widetilde{\bbQ} \widetilde{\bbR}$, where $\widetilde{\bbQ}$ is a matrix with orthogonal columns and $\widetilde{\bbR}$ is upper triangular with ones in the diagonal. This decomposition is unique. Indeed, one may obtain $\widetilde{\bbQ}$ by a Gram-Schmidt process, but skipping the normalization steps. A more efficient approach is to consider the classical QR decomposition ($\bbZ=\bbQ\bbR$, with $\bbQ$ orthonormal and $\bbR$ upper triangular), and compute $\widetilde{\bbQ} = \bbQ \bbD_R$, where $\bbD_R=\textrm{diag}(\bbR)$ is the diagonal matrix with the diagonal entries of $\bbR$. In a way, this modification of the QR decomposition shifts the ``normalization'' of the columns from the upper triangular factor towards the orthogonal factor.
	
	Note that $\widetilde{\bbQ}\in \mathcal{M}$ and this decomposition will serve to define a retraction to the manifold in the next proposition. {Again, this procedure differs from the popular $Q$-factor retraction to the Stiefel manifold~\cite[p. 160]{boumal2023intromanifolds}.}
	
	\begin{myproposition}\label{prop:retraction}
		Let $\bbX\in\mathcal{M}$ and $\bbzeta \in T_{\bbX}\mathcal{M}$ a tangent vector. Then, the mapping 
		\begin{equation*}
			R_{\bbX}(\bbzeta) = \widetilde{qf}(\bbX+\bbzeta)
		\end{equation*}
		is a retraction, where $\widetilde{qf}(\bbA)$ denotes the $\widetilde{\bbQ}$ factor of the modified QR decomposition described above, and the sum $\bbX+\bbzeta$ stands for the usual abuse of notation for embedded manifolds on vector spaces.
	\end{myproposition}
	
	We now have all the ingredients for the GD method to minimize $f(\bbX^l,\bbX^r)=\|\bbM\circ(\bbA-\bbX^l(\bbX^r)^\top)\|_F^2$ over $\mathcal{M}$, which is tabulated under Algorithm \ref{algo:mani_grad_descent}.
	The convergence rate of Riemannian GD is the same as the unconstrained counterpart (i.e., producing points with $\grad f$ smaller than $\varepsilon$ in $\mathcal{O}(1/\varepsilon^2)$ iterations) \cite{boumalAbsil}. The computational complexity of each iteration is dominated by the QR decomposition in the retraction.
	
	\begin{algorithm}[t]
		\caption{Riemannian Gradient Descent (GD) on $\ccalM$}
		\label{algo:mani_grad_descent}
		\algsetup{linenosize=\normalsize}
		\begin{algorithmic}[1]
			\REQUIRE Initial $\bbX^l_0$ and $\bbX^r_0$
			\REPEAT
			\STATE Compute Euclidean gradients\\ $\nabla f_{\bbX^l}(\bbX^l_k,\bbX^r_k)$ and $\nabla f_{\bbX^r}(\bbX^l_k,\bbX^r_k)$
			\STATE Compute Riemannian gradients as \\$\grad f_{\bbX^l}(\bbX^l_k,\bbX^r_k) = \pi_{\bbX^l_k}(\nabla f_{\bbX^l}(\bbX^l_k,\bbX^r_k))$\\
			$\grad f_{\bbX^r}(\bbX^l_k,\bbX^r_k) = \pi_{\bbX^r_k}(\nabla f_{\bbX^r}(\bbX^l_k,\bbX^r_k))$
			\STATE Compute retraction with $\alpha$ chosen via the Armijo rule\\$\bbX^l_{k+1} = R_{\bbX^l_k}\left( {- \alpha \grad f_{\bbX^l}(\bbX^l_k,\bbX^r_k)}\right)$,\\
			$\bbX^r_{k+1} = R_{\bbX^r_k}\left( {-\alpha \grad f_{\bbX^r}(\bbX^l_k,\bbX^r_k)}\right)$
			\UNTIL{convergence}
			\RETURN $\{\bbX^l_k,\bbX^r_k\}$.
		\end{algorithmic}
	\end{algorithm}
	

	We extended the \texttt{Pymanopt} package \cite{pymanopt} with a class for the manifold $\ccalM$ defined in Proposition \ref{stiefel_tilde_manifold}, which forms part of the code available for this paper.
	
	\begin{remark}[Rescaling the factors' columns]\label{rem:scaling_cols}\normalfont Algorithm \ref{algo:mani_grad_descent} does not quite solve \eqref{eq:dase_mask_constrain}. While both $\{\bbX^l_k,\bbX^r_k\}$ belong to $\ccalM$, the constraint $(\bbX_k^l)^\top\bbX_k^l = (\bbX_k^r)^\top\bbX_k^r$ will in general not be satisfied upon convergence. Dropping the iteration index for simplicity, let $\bar{\bbx}_i^l, \bar{\bbx}_i^r\in\reals^{N}$ be the $i$-th columns of $\bbX^l$ and $\bbX^r$, respectively. To obtain a feasible solution from the output of Algorithm \ref{algo:mani_grad_descent}, for each dimension $i=1,\ldots,d$ we define scaling factors $s_i=\|\bar{\bbx}_i^l\|_2/\|\bar{\bbx}_i^r\|_2$ and collect them in the diagonal matrix $\bbS=\textrm{diag}(s_1,\ldots,s_d)$. We then rescale the columns of the embedding matrices via $\bbX^l_k \leftarrow \bbX^l_k\bbS^{-1/2}$ and $\bbX^r_k\leftarrow \bbX^r_k\bbS^{1/2}$, without affecting the value of the objective function but now satisfying the constraint in \eqref{eq:dase_mask_constrain}.
	\end{remark}
	
\begin{figure}[t]
		\centering
		\includegraphics[width=\linewidth]{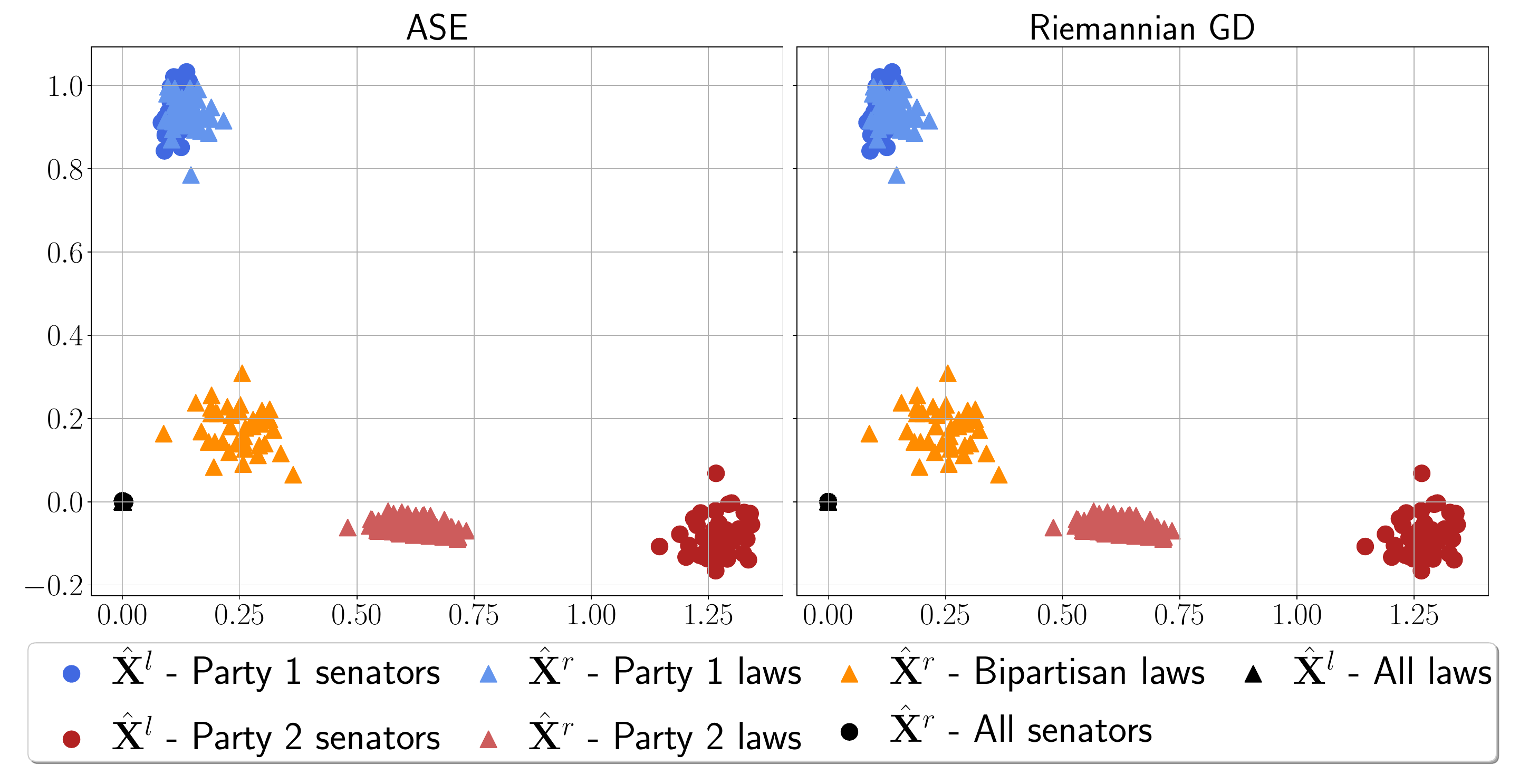}
		\caption{Solution to the embedding problem \eqref{eq:dase_mask_constrain} for the bipartisan senate example. ASE (left) and Riemannian GD (right). Notice how both solutions are nearly identical [cf. unconstrained GD in Fig. \ref{fig:embeddings_bipartidismo_no_ortogonal} (right)], underscoring the importance of the orthogonality constraints. }\label{fig:embeddings_bipartidismo_manifold}
	\end{figure}

	\begin{myexample}[Bipartisan senate revisited]\label{Ex:senate_revisited}\normalfont
		Going back to the bipartisan senate from Example \ref{Ex:senate}, Fig. \ref{fig:embeddings_bipartidismo_manifold} depicts the solution of \eqref{eq:dase_mask_constrain} for the same simulated bipartite senator-law digraph (imposing the orthogonality constraints and rescaling in Remark \ref{rem:scaling_cols}). Unlike in Example \ref{Ex:senate}, the Riemannian GD algorithm on the manifold $\ccalM$ is able to recover the same structure as the ASE. Laws are now correctly aligned with their corresponding party, thus faithfully revealing the structure in the data.
	\end{myexample}
	
	\section{Numerical Experiments and Applications}\label{sec:applications}
	
	In this section we illustrate our embedding algorithms' ability to produce accurate and informative estimates of nodal latent position vectors. We explore a variety of GRL  applications and consider synthetic and real network data. Our test cases are designed to target ASE challenges outlined in Section \ref{subsec:ase_challenges}, namely: i) missing data; ii) embedding multiple networks; and iii) graph streams (with fixed and varying number of nodes). For each case we assess the results with respect to estimation accuracy, interpretability, and stability/alignment in dynamic environments. {The suitability of spectral embeddings (rooted in the RDPG
		model) for downstream tasks has already been well-documented~\cite{oos2018levin,priebe2018survey}. For this reason, the goal here is to demonstrate the effectiveness of our algorithms in generating
		node embeddings that faithfully represent network structure in novel settings i)-iii). Supplementary results exploring algorithm sensitivity to random initialization are in Appendix \ref{sec:app_robust}.} The code for all these experiments is publicly available at \url{https://github.com/marfiori/efficient-ASE}.

	
	
	\subsection{Inference with missing data}\label{sec:missing_data}
	
	
	First we illustrate how GD-based inference can be useful for GRL with missing data. 
	The setup is similar to that of Example \ref{Ex:senate}, but here we rely on real United Nations (UN) General Assembly voting data~\cite{DVN/LEJUQZ_2009}. For each roll call and country, the dataset includes if the country was present and if so the corresponding vote (either `Yes', `No', or `Abstain') for each proposal. We analyze the associated bipartite digraph pertaining to a particular year, where nodes correspond to countries and roll calls, and an edge from a country to a roll call exists if it voted affirmatively. If the country was absent or abstained, we will tag that edge as unknown ($M_{ij}=0$).  
	
\begin{figure}[t!]
		\centering
		\includegraphics[width=\linewidth]{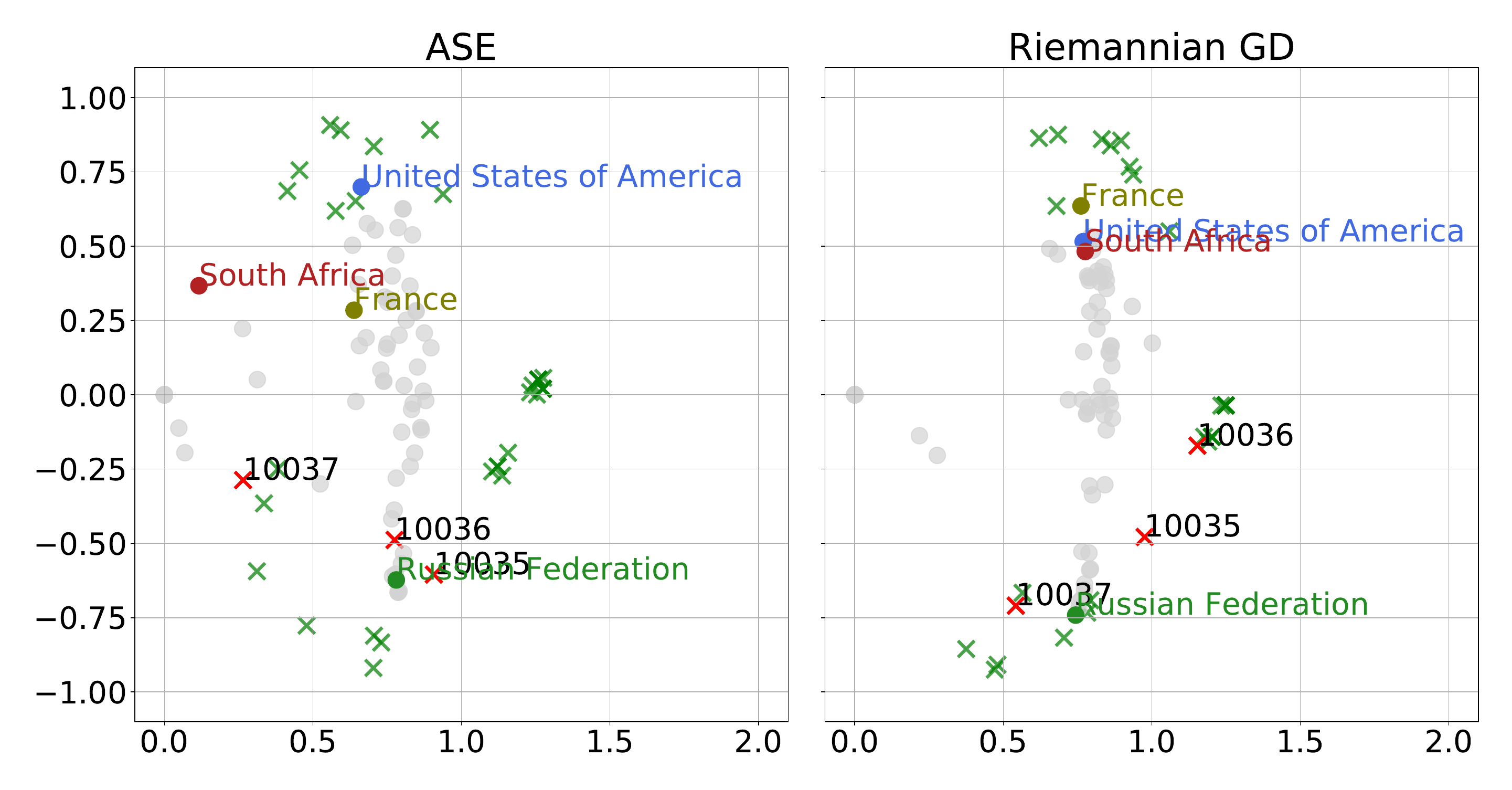}
		\caption{UN General Assembly voting data for 1955. ASE (left) and Riemannian GD {(i.e.,\ Algorithm \ref{algo:mani_grad_descent})} with mask matrix encoding present and absent (or abstained) voters (right). Our approach is able to assign the absent voters to the correct group (e.g., South Africa) and offers a more clear clustering of roll calls.}
		\label{fig:embeddings_onu_1955}
	\end{figure}

	Fig. \ref{fig:embeddings_onu_1955} depicts the node embeddings ($d=2$) of the graph from 1955, estimated by ASE (naively assuming unknown edges do not exist, $A_{ij}=0$) and Riemannian GD {(i.e.\ Algorithm \ref{algo:mani_grad_descent})}. Consider the countries, which are displayed as circles. We highlight four interesting cases: Russia, USA, France, and South Africa. At the time, the first two represented two poles of the world, and are naturally almost orthogonal to each other for both methods. Note furthermore how the ASE seems to indicate that South Africa is less likely to vote in agreement with Russia than (even) the USA, whereas the opposite is true for France. 
	
	The problem comes from equating an absence or abstention to a negative vote. For instance, South Africa was only present in roughly one third of the roll calls, and it voted almost identically to the USA. The Riemannian GD method, which acknowledges unknown edges via the mask $\bbM$, provides an embedding that reflects this agreement. Something similar happens with France, which differed from the USA only in six roll calls. Four correspond to USA abstentions and France voting `Yes', another one where the opposite happened (and thus both cases should not be accounted for in the optimization problem), and finally the roll call 10036 where France was one of only two countries to vote `No' (the USA voted `Yes'). 
	
	Regarding the embeddings of roll calls marked with a cross in Fig.\ \ref{fig:embeddings_onu_1955}, note how 10036 is aligned with the Russian block of countries by ASE, but it is better placed as an intermediate proposal in Fig.\ \ref{fig:embeddings_onu_1955} (right) -- equally likely to be voted by all countries. Something similar occurs with roll call 10035, which dealt with the same subject of 10036, but met resistance from more countries (roughly a dozen, including the USA and France). In both cases several countries were not present or abstained during the voting. Incorrectly assuming these votes as negative by ASE leads to biased results. Much more can be said about the roll calls and their associated UN resolutions, but let us conclude the discussion by noting that roll call embeddings generated by Algorithm \ref{algo:mani_grad_descent} form three clusters reflecting the geopolitical landscape at the time. There is a cluster for each pole (American and Russian), plus an intermediate one where both poles tend to vote similarly. On the other hand, ASE generates roll call embeddings that are incorrectly aligned (e.g., 10036), and a loose grouping of intermediate roll calls with shared voting from both poles. 
	
	\subsection{Embedding multiple graphs: the batch case}\label{sec:warm}
	
	
	Suppose now that we observe $m>1$ graphs $\{\bbA_t\}_{t=1}^m$ and we are interested, for instance, in testing whether they are drawn from the same RDPG model, or, in tracking the embeddings over time. Assume that we can identify nodes across different observations; e.g., they correspond to labeled users in a social network and so a matching algorithm is not needed. Independently obtaining the ASE for each graph is undesirable because it yields arbitrarily rotated embeddings, a challenge that has motivated several recent research efforts. 
	
	Indeed, a hypothesis test which involves solving a Procrustes problem to align the embeddings was put forth in~\cite{tang2017semiparametric}. An alignment alternative is to jointly embed all $m$ graphs via a single `super-matrix' decomposition. The so-called \emph{Omnibus} embedding first forms an $mN \times mN$ matrix derived from all $\{\bbA_t\}_{t=1}^m$, and then computes its ASE which enjoys asymptotic normality~\cite{levin2017omnibus}. The Unfolded ASE (UASE) also constructs an auxiliary matrix, but by horizontally stacking all $\{\bbA_t\}_{t=1}^m$~\cite{jones2020multilayer,gallagher2021spectral}. Nodal representations are then extracted from the SVD of this $N\times mN$ matrix. Under some technical assumptions, the UASE provably offers desirable longitudinal and cross-sectional stability~\cite{gallagher2021spectral}. However, the complexity and memory footprint of these \emph{batch} approaches grow linearly with $m$, and they are only applicable to undirected graphs.
	
	In the context of the algorithms proposed in this paper, we may leverage their iterative nature and initialize them using the estimated embeddings of another related (e.g., contiguous in time) graph. Unless radical changes take place from one graph to the other, this so-called warm restart is expected to produce embeddings that are closely aligned, with the added benefit of converging in few iterations.\vspace{2pt}
	
	\noindent \textbf{Stability of {BCD} estimates.} Let us illustrate this (desirable) behaviour through a numerical example. We borrow the setting and code from~\cite{gallagher2021spectral}. Consider two graph samples drawn from a dynamic SBM with inter-community probability matrices
	$$
	\bbPi_1 = \begin{psmallmatrix}
		0.08 & 0.02 & 0.18 & 0.10 \\
		0.02 & 0.20 & 0.04 & 0.10 \\
		0.18 & 0.04 & 0.02 & 0.02 \\
		0.10 & 0.10 & 0.02 & 0.06
	\end{psmallmatrix},\:
	\bbPi_2 = \begin{psmallmatrix}
		0.16 & 0.16 & 0.04 & 0.10 \\
		0.16 & 0.16 & 0.04 & 0.10 \\
		0.04 & 0.04 & 0.09 & 0.02 \\
		0.10 & 0.10 & 0.02 & 0.06
	\end{psmallmatrix}.
	\label{eq:uase_experiment}
	$$ 
	Initially there are four communities. At time $2$, the first two communities merge, community $3$ moves, and community $4$ has its connection probabilities unchanged. 
	
\begin{figure}[t!]
		\centering
		\includegraphics[width=0.46\textwidth]{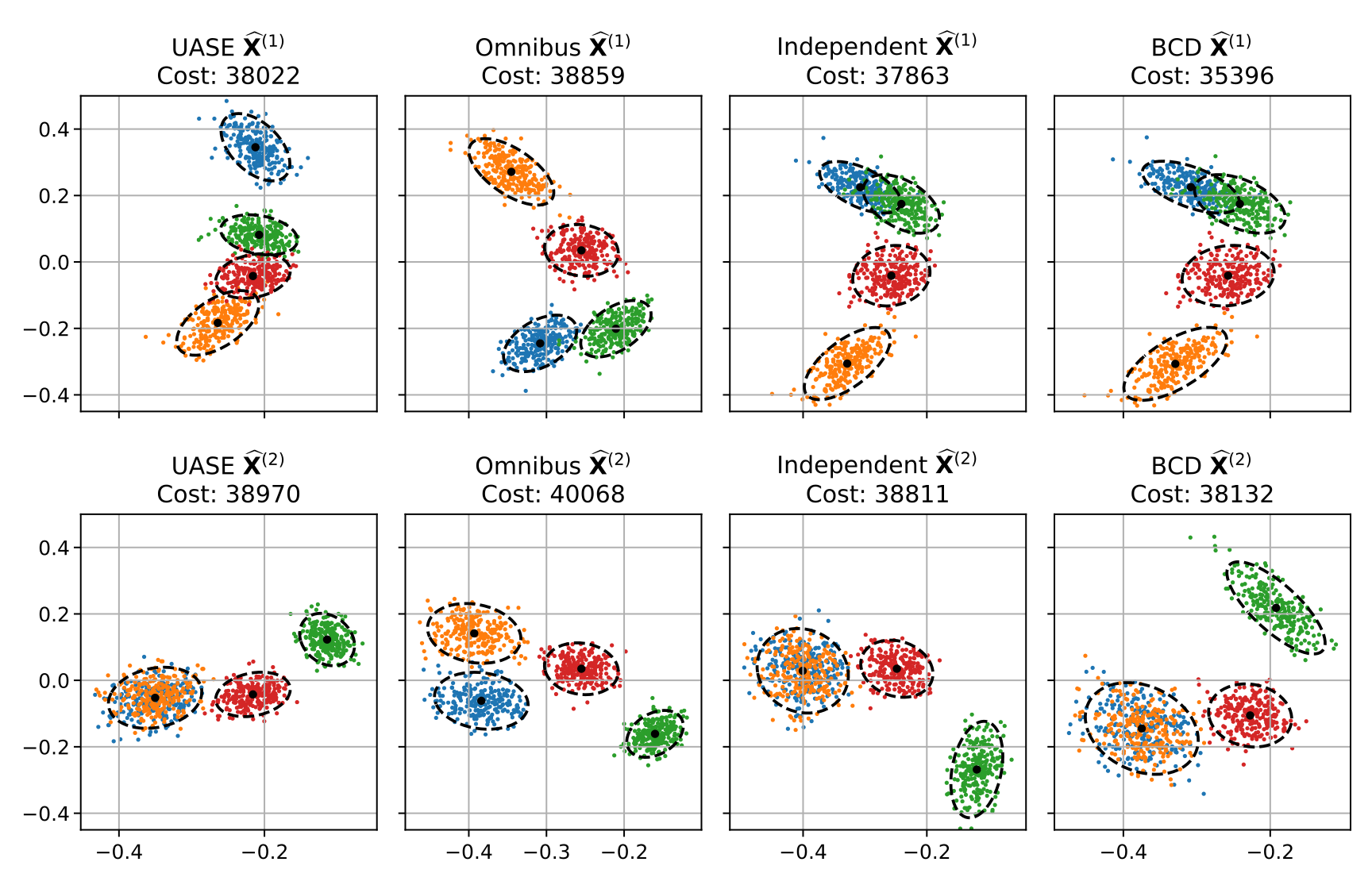}
		\caption{Embeddings of two SBM graph realizations, where communities $1$ and $2$ merge, while community $4$ keeps the connection probabilities with other groups. Observe how the {BCD} approach (far right) manages to capture this behaviour, while providing the best representation for each graph individually (quantified by the smallest cost function values). Example adapted from \cite{gallagher2021spectral}.}
		\label{fig:vs_UASE}
	\end{figure}

	Ideally, when embedding both graphs: i) the representations of nodes in community $4$ should not change (longitudinal stability); and ii) the time $2$ embeddings of members of communities $1$ and $2$ should be similar, up to noise (cross-sectional stability). Fig.~\ref{fig:vs_UASE} displays the results for UASE~\cite{gallagher2021spectral}, Omnibus embedding~\cite{levin2017omnibus}, independent ASE for each graph, and {BCD} ({i.e.\ Algorithm \ref{algo:coord_descent}} warm-restarted at time $2$ {with the result of time $1$}). 
	As expected, independent ASE lacks longitudinal stability, and the Omnibus embedding fails to exhibit cross-sectional stability. Note how the time 2 Omnibus estimates of communities 1 and 2 remain separate, due to time 1 `interference' affecting this joint embedding.
	
	UASE and {BCD} produce embeddings that fulfill both stability requirements i) and ii). However, {BCD} yields a better overall representation for both graphs. 
	This is quantified via the cost function \eqref{eq:ase_mask} evaluated at each solution; see above each plot for the numerical values. Unlike the batch UASE, {gradient descent methods as the ones we present here offer} a pathway towards tracking nodal representations in a streaming graph setting -- the subject dealt with next. 
	
	
	\subsection{Model tracking for graph streams}\label{sec:tracking}
	
	Consider now a monitoring scenario, where we observe a stream of time-indexed graphs $\{\bbA_t\}$ and the goal is to track the underlying model. Different from the batch setting of the previous section, we are now unable to jointly process the entire graph sequence. This may be due to memory constraints or stringent delay requirements. 
	We will still assume that nodes are identifiable across time, but the algorithm's computational cost and memory footprint may not increase with $t$. 
	
	\subsubsection{Fixed vertex set}
	
	We first consider the setting where $N$ is fixed and we would like to track the latent vectors $\bbX_t\in\reals^{N\times d}$.\footnote{We stick to undirected graphs for ease of exposition, but extensions to digraphs are straightforward and presented in the numerical experiments.}  
	Previous efforts in this direction have been mainly motivated by the change-point detection problem; i.e., detecting if and when the generative model of the observed graph sequence changes~\cite{yu2021onlineCPD,marenco2021tsipn,chen2019sequentialCPD,zhang2020icassp}. 
	Our focus is on the related problem of estimating the  embeddings' evolution. A couple noteworthy applications include recommender systems (where rankings are revealed, or even change, over time)~\cite{campos2014time} or, as we discuss below, monitoring wireless networks~\cite{mateos2013cartography}. 
	
	
	Independent ASE computation for each $\bbA_t$ suffers from the alignment issue already discussed. 
	Instead, and supposing for now that $\bbM$ can be ignored, 
	it may well be the case that recursive methods to update the SVD of a perturbed matrix $\bbA_t=\bbA_{t-1}+\mathbf{\Delta}_t$ suffice~\cite{brand2006fast}. However, as we show in the following synthetic example, these approaches may also produce arbitrarily rotated estimates from one time-step to the next, and suffer from catastrophic error accumulation~\cite{zhang2018timers}. 
	
\begin{figure}
		\centering
		\includegraphics[width=0.6\linewidth]{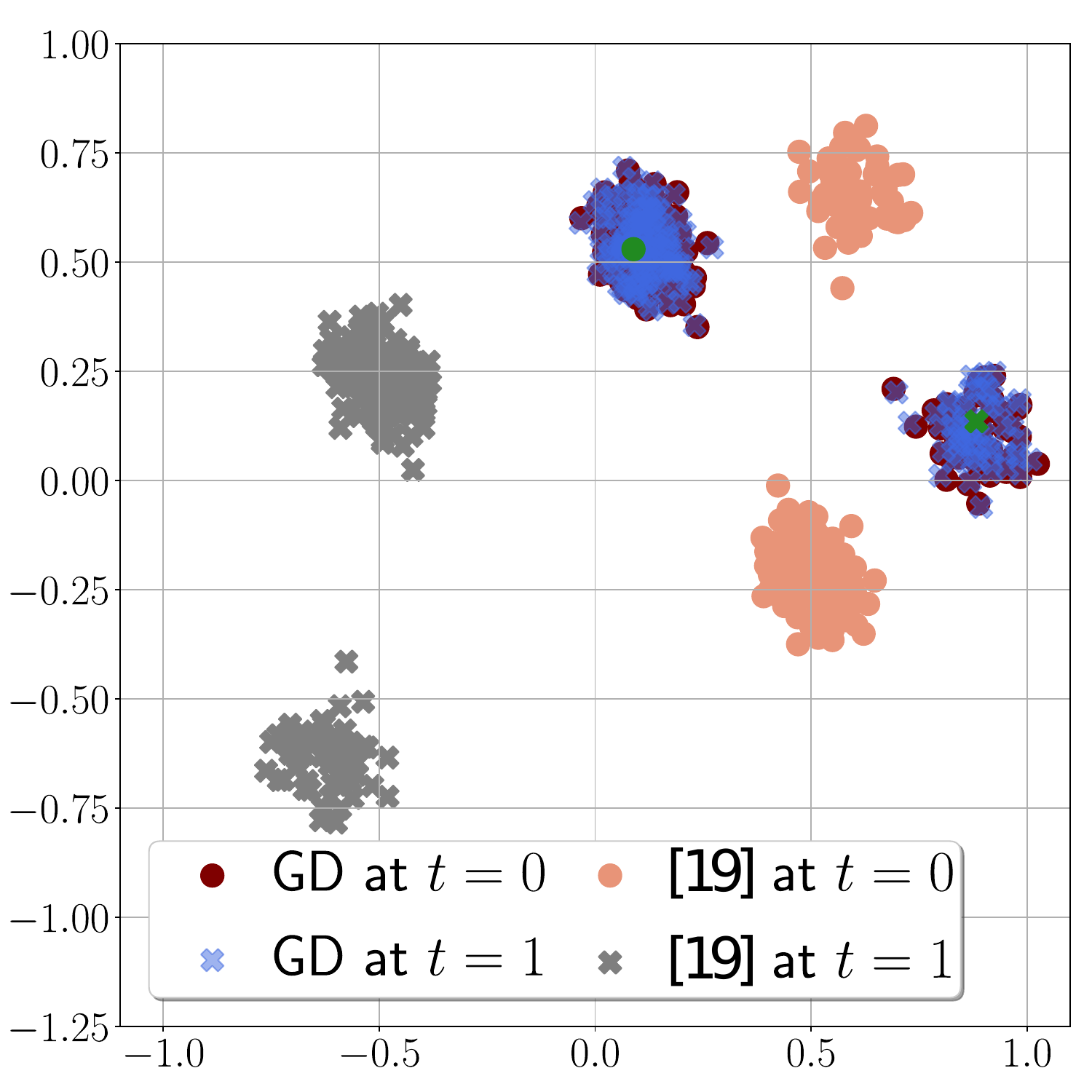}
		\includegraphics[width=\linewidth]{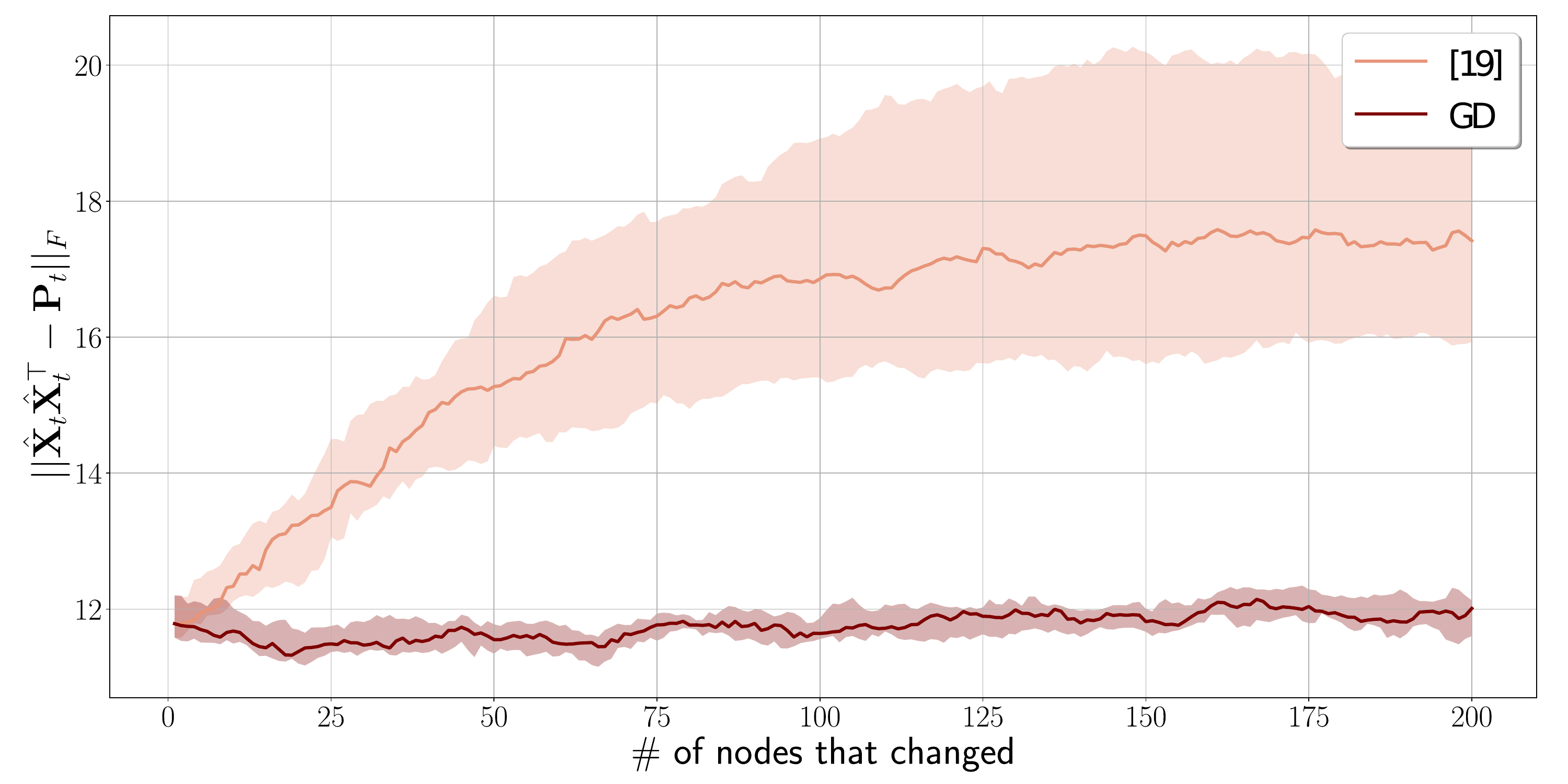}
		\caption{Two-block dynamic SBM in which a single node changes affiliation at each $t$. Comparison between online GD and recursive SVD~\cite{brand2006fast}.  (top) Embeddings for the first two time-steps ($d=2$); the node that changed communities is highlighted in green. Best viewed in a color display. Note how the change of a single node produces markedly different results for~\cite{brand2006fast}, whereas online GD offers stable estimates. (bottom) Evolution of $\|\hbX_t\hbX_t^\top-\bbP_t\|_F$. Solid line indicates median across ten realizations, with the range between first and third quartiles shown in a lighter color. Online GD exhibits uniformly bounded error, whereas~\cite{brand2006fast} accumulates error as $t$ grows.} 
		\label{fig:rotation}
	\end{figure}

	\vspace{2pt} 
	{\noindent \textbf{Tracking of a dynamic SBM.} }
	Our idea is instead to proceed as in Remark \ref{rem:warm_restarts}, and warm-restart the GD iterations with the previous time-step's estimate $\hbX_{t-1}$ {(analogously to the example in Fig. \ref{fig:vs_UASE})}. 
	Consider a dynamic SBM graph with $N=200$ nodes and two communities. At each time-step $t=0,1,2,\ldots$ a single randomly chosen node changes its community affiliation. We compare the tracking performance of warm-restarted GD {[i.e.,\ several iterations of GD in \eqref{eq:gd} initialized with the previous time-step's estimate]} and the fast, recursive SVD algorithm in~\cite{brand2006fast}. The nodal embeddings for $t=0$ and $1$ (i.e., a single node changed affiliation) are depicted in Fig.\ \ref{fig:rotation} (top). Notice how online GD produces stable results, with a single vector moving from one cluster to the other. The rest of the nodes' embeddings remain virtually unchanged. On the other hand, the recursive SVD in~\cite{brand2006fast} fails to preserve a common angular reference for $\hbX_0$ and $\hbX_1$. Another well-documented drawback of these incremental SVD methods is that, since they update only the $d$ most significant components, the error $\|\hbX_t\hbX_t^\top-\bbP_t\|_F$ increases with $t$~\cite{zhang2018timers}. Fig. \ref{fig:rotation} (bottom) illustrates this error-accumulation behavior, to be contrasted with online GD that keeps the error in check for all $t\geq 0$.\vspace{2pt}

	
	{\noindent \textbf{Wireless network monitoring.} }
	We may further leverage the fact that $\bbX_t$ is typically correlated with $\bbX_{t-1}$ in order to improve the embeddings' accuracy over time.
	For example, suppose $\bbX_{t-m}=\ldots=\bbX_{t}=\bbX$ over some interval of length $m$. It is then prudent to estimate $\bbX$ by solving~\eqref{eq:ase_mask}, but using the average $\barbA_t=1/m\sum_{k=t-m}^{t}\bbA_k$ instead~\cite{tang2017robust}. 
	Note that $\barbA_t$ may be interpreted as the adjacency matrix of a weighted graph. Edge weights can also be modeled by an RDPG, where now the embeddings are such that $\bbX\bbX^\top = \E{\bbA}$. Unlike the unweighted case, $\E{\bbA}$ are not probabilities. Still, under mild assumptions on the weights' distribution, the solution of \eqref{eq:ase_mask} for weighted $\bbA$ is a consistent estimator of $\bbX$ as $N\to \infty$~\cite{marenco2021tsipn}.
	
	\begin{figure}
		\centering
		\includegraphics[width=0.45\textwidth]{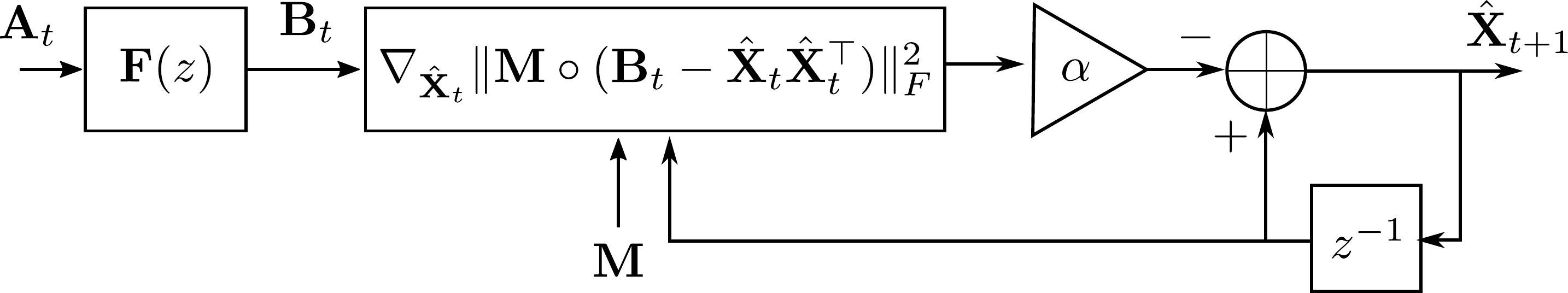}
		\caption{A diagram of the proposed tracking system. The entry-wise filter $\bbF(z)$ implements an averaging operator, e.g., a fixed-length moving average. 
		}
		\label{fig:diagrama_tracking}
	\end{figure}
	
	These observations motivate well the two-stage tracking system depicted in Fig.\ \ref{fig:diagrama_tracking}. The stream of adjacency matrices $\{\bbA_t\}$ is fed to the entry-wise filter $\bbF(z)$, which outputs $\{\bbB_t\}$. For instance, $\bbF(z)$ may be a moving average of fixed length $m$ as before. If memory is at a premium, we may use a single-pole IIR filter instead so that $\{\bbB_t\}$ is an exponentially-weighted moving average of the input adjacency matrices.  We may even drop the filtering stage altogether (setting $m=1$) to yield a least mean squares (LMS)-type online GD algorithm. 
	
	We now empirically demonstrate this simple tracking system yields accurate and stable embeddings of dynamic RDPG graphs.
	Consider a Wi-Fi network from which a monitoring system periodically acquires the Received Signal Strength Indicator (RSSI) between Access Points (APs) -- a feature typically available in enterprise-level deployments. We will use our GRL framework to flag network changes and eventually diagnose them. We analyze graphs $\bbA_t$ whose nodes are the APs and the edge weights are the measured RSSI values (plus a constant offset so that all values are positive). Since these measurements are typically not symmetric, we have a digraph sequence. We rely on the dataset described in~\cite{capdehourat2020nation}, which consists of hourly measurements between $N=6$ APs at a Uruguayan school, over almost four weeks ($m=655$ graphs). During the monitoring period, the network administrator moved an AP ($i=4$) at $t\approx 310$. 
	
	To track the AP embeddings, {we run an online version of Algorithm \ref{algo:mani_grad_descent} as schematically shown in the diagram of Fig. \ref{fig:diagrama_tracking}, but adapted to digraphs}. This entails a retraction after the Riemannian GD step, not shown in the diagram.
	We use an IIR filter $\bbF(z)$ with a pole at $0.9$.  Furthermore, we adopt a fix stepsize $\alpha=0.01$ instead of choosing it via the Armijo rule. 
	
	
	\begin{figure}[t]
		\centering
		\includegraphics[width=\linewidth]{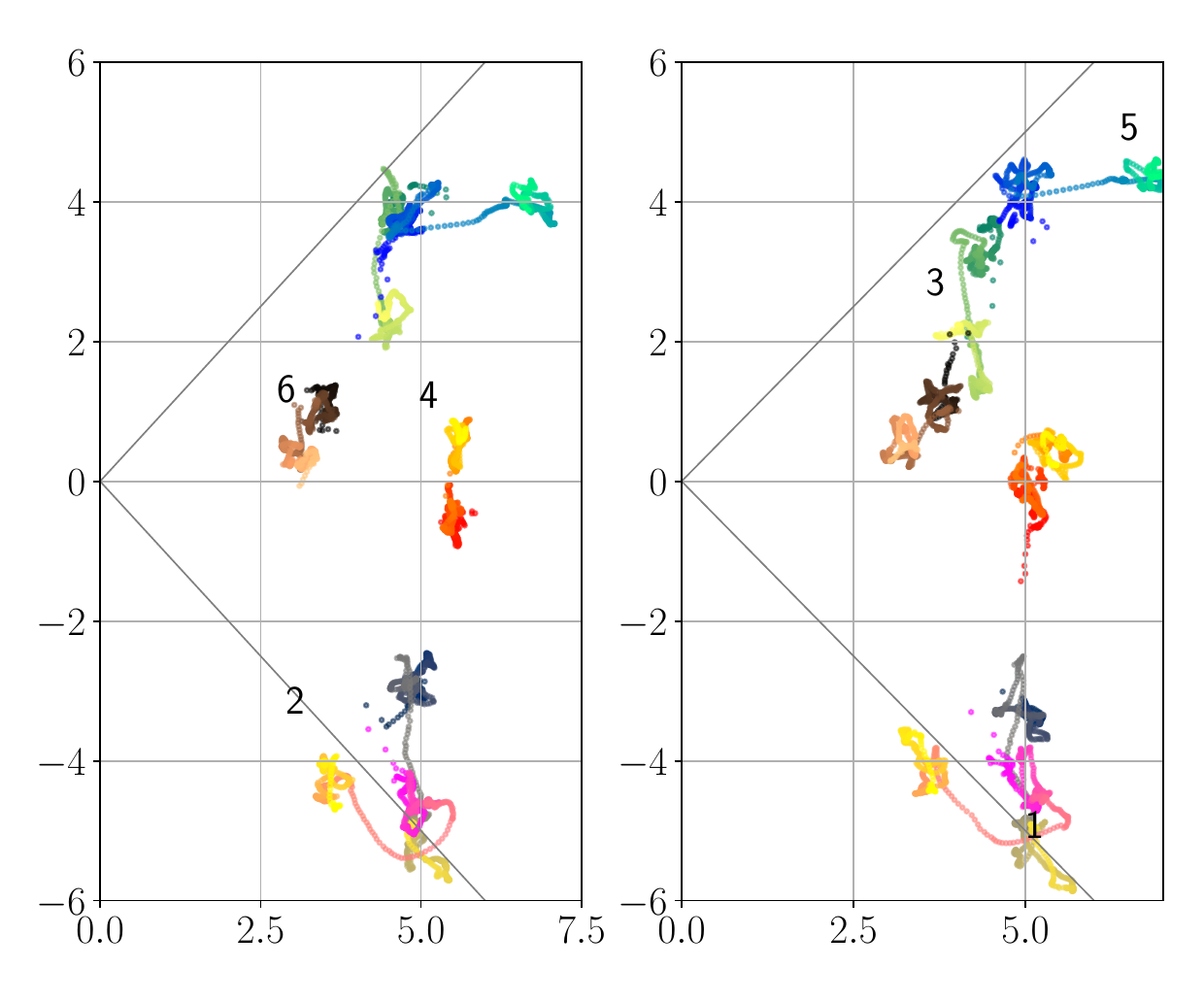}
		\caption{Embeddings $\hbX^l_t$ (left) and $\hbX^r_t$ (right) for the RSSI digraph ($d=2$). Color palettes distinguish the APs and a lighter tone indicates larger values of $t$. Best viewed in a color display. The network's change at $t\approx 310$ is apparent. AP $4$ was moved ($i=4$) closer to the upper cluster of APs. }
		\label{fig:rssi}
	\end{figure}
	
	The evolution of the {online Riemannian GD} estimates $\hbX^l_t$ and $\hbX^r_t$ for $d=2$ is shown in Fig. \ref{fig:rssi}. Different color palettes are used to distinguish the nodes, and as $t$ increases the colors become lighter. Note how at all times there are two (almost) orthogonal cluster of nodes: c1) APs $1$ and $2$ (in the lower part of the plots); and c2) APs $3$, $5$ and (to a lesser extent) $6$. AP $4$ is embedded between both communities for all $t$. Moreover, note how the \emph{trajectory of each AP} can be split into a couple clear states, discernable as the colors transition from darker to lighter. This is indicative of the change in AP $4$'s position at roughly the middle of the monitoring period. Finally, movement within both AP clusters is mostly radial, hence dot products between cluster members are preserved. On the other hand, AP $4$ moves tranversally closer to c2, consistent with the information provided by the network administrator. It appears as if it was moved closer to AP $5$, and AP $1$ remains its closest member from c1. 
	
	\subsubsection{Time-varying node set}
	
	In dynamic environments it is not uncommon for nodes to join or leave the network. Going back to the wireless network test case, the question remains on how to proceed should an AP fail, or if the administrator decides to add a new one to improve coverage. Dealing with the former case is straightforward; if a node leaves the network at time $t$, we simply drop the corresponding row in $\hbX_{t-1}$ and re-run the GD algorithm (warm-restarted from there). 
	
	Node additions require more thought. Suppose that a single node $i=N+1$ joins the network at time $t$. Let $\bba_{N+1}=[A_{1,N+1},\ldots,A_{N,N+1}]^\top\in\{0,1\}^N$ be the $(N+1)$-th column of $\bbA_t\in\{0,1\}^{N+1\times N+1}$, excluding $A_{N+1,N+1}=0$ and dropping the subindex $t$ for notational convenience. Then given $\hbX_{t-1}\in\reals^{N\times d}$, we can embed node $i$ by solving
	\begin{equation}\label{eq:projection}
		\hbx_{N+1}=\argmin_{\bbtheta\in\reals^d}\|\bba_{N+1}-\hbX_{t-1}\bbtheta\|_2^2.   
	\end{equation}
	%
	%
	%
	%
	This simple but intuitive out-of-sample embedding procedure was studied in~\cite{oos2018levin}, and shown to recover the true latent positions as $N\to\infty$. If several nodes are added at a given time-step, they can all be embedded by solving multiple LS problems like~\eqref{eq:projection}. However, this procedure disregards the information from the connections between new nodes. Furthermore, if the embeddings of existing nodes are not updated, their growing inaccuracies as $\bbA_t$ evolves will negatively impact future nodes' representations. 
	
	As we show in the following numerical experiments, these drawbacks can be overcome by running our online GD-based algorithms to update \emph{all embeddings} $\hbX_t$, initializing existing nodes with $\hbX_{t-1}$ and new one(s) with $\hbx_{N+1}$ as in ~\eqref{eq:projection}.\vspace{2pt}
	
\begin{figure}[t]
		\centering
		\includegraphics[width=\linewidth]{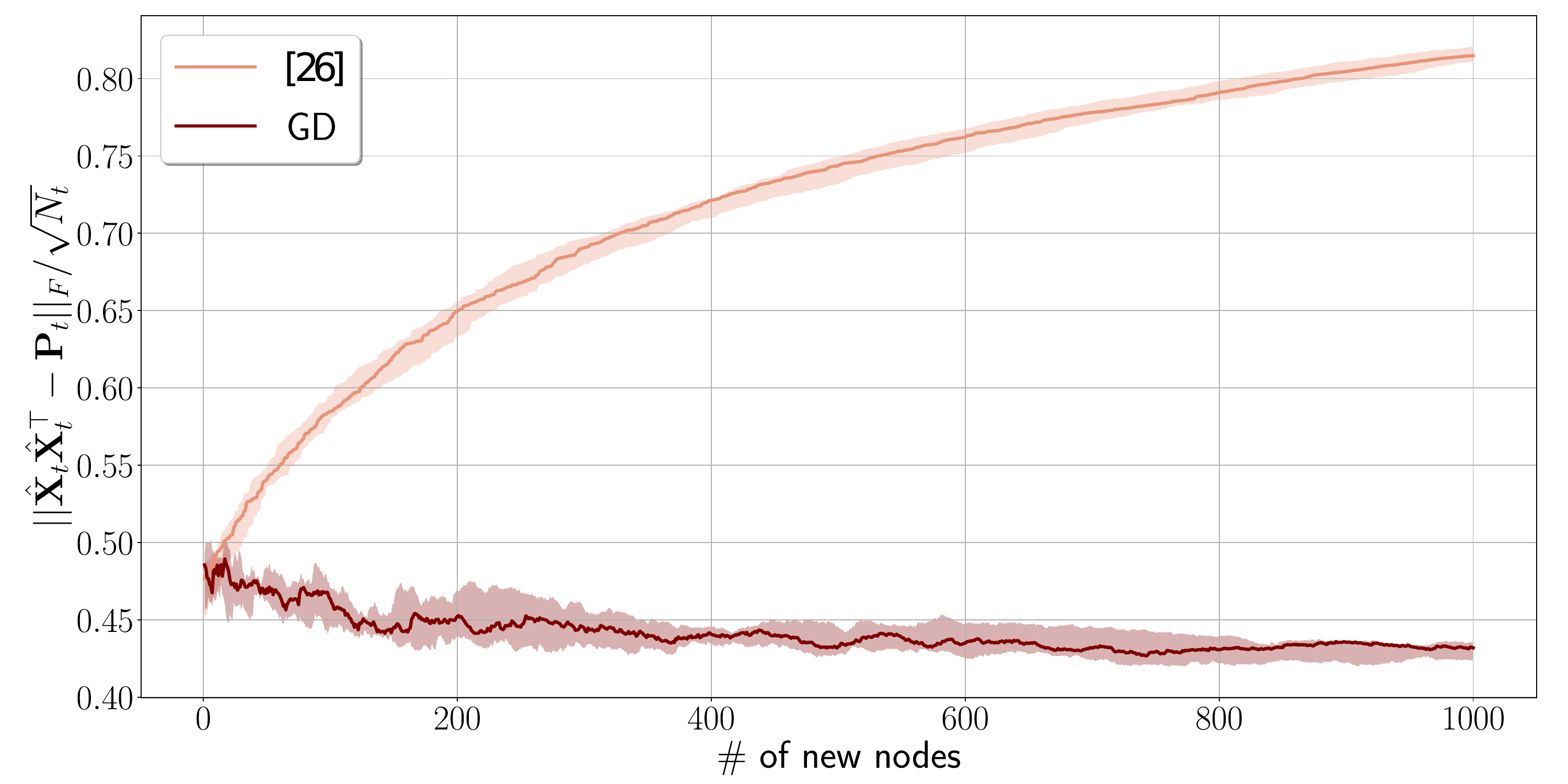}
		\caption{Dynamic Erd\"os-R\'enyi graph in which a single node is added at each $t$. Comparison between online GD and out-of-sample LS embedding~\cite{oos2018levin}. Evolution of $\|\hbX_t\hbX_t^\top-\bbP_t\|_F/\sqrt{N_t}$. Solid line indicates median across ten realizations, with range between first and third quartiles shown in a lighter color. Once more, online GD exhibits uniformly bounded error, whereas the baseline method~\cite{oos2018levin} accumulates error as $t$ grows.}
		\label{fig:mse_priebe}
	\end{figure}

	
	\noindent\textbf{Dynamic random graph with growing vertex set.} Consider an Erd\"os-R\'enyi graph with a fixed connection probability $p=0.1$, and initial number of nodes $N_0=100$. At each time-step $t$ we add a single node so that $N_{t}=N_{t-1}+1$. The evolution of the error $\|\hbX_t\hbX_t^\top-\bbP_t\|_F/\sqrt{N_t}$ is shown in Fig. \ref{fig:mse_priebe}. Note how (carefully warm-restarted) {online GD} exhibits bounded error behavior, in stark contrast with repeated LS-based embeddings as in~\cite{oos2018levin}. Admittedly, this gain in accuracy comes with a modest increase in computation (few  GD steps), and identical memory footprint (i.e., storing the current embeddings and the new adjacency matrix) as the baseline in~\cite{oos2018levin}.\vspace{2pt}
	
	
	
	\noindent\textbf{Tracking international relations from UN voting data.} Here we revisit the UN General Assembly voting data from Section \ref{sec:missing_data}. Following the same bipartite digraph construction procedure, we study all yearly graphs from 1955 to 2015. 
	In this dynamic network we have a time-varying node set. Roll calls change from one year to the next, and also several countries joined the UN later (while others have ceased to exist). We embed the first graph from 1955 using Riemannian GD initialized at random (as before, using $d=2$). For each successive year, we warm-restart {Algorithm \ref{algo:mani_grad_descent}} with the embeddings from the previous year, while new nodes are initialized using the LS solution~\eqref{eq:projection}.
	
\begin{figure}[t]
		\centering
		\includegraphics[width=0.7\linewidth]{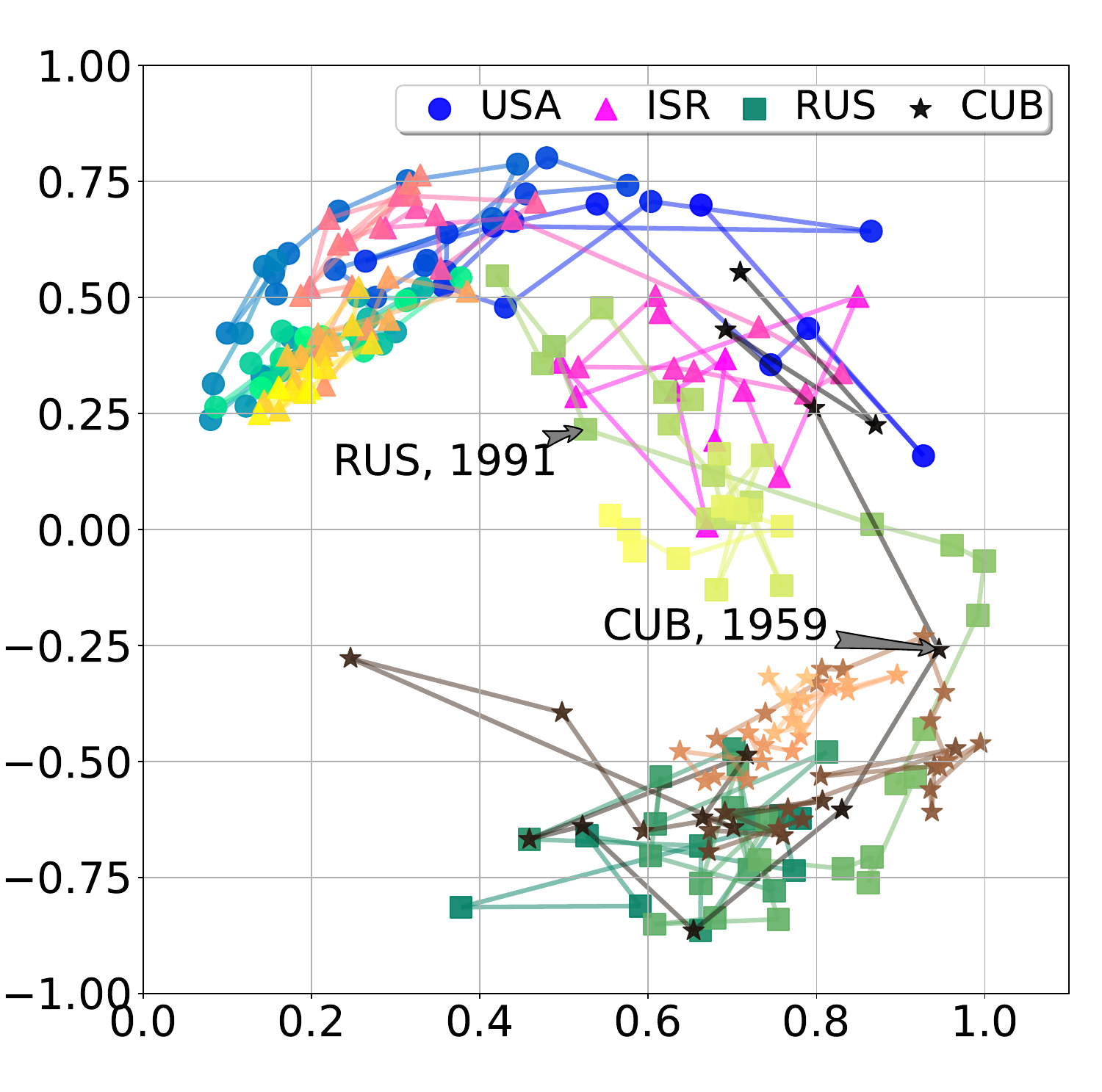}
		\caption{UN General Assembly voting data from 1955 to 2015. Evolution of nodal positions for the USA, Israel, Cuba, and the USSR (or, after 1991, the Russian Federation) estimated via online Riemannian GD. Color palettes distinguish the countries and a lighter tone indicates later years. Best viewed in a color display. Note how the USA and Israel remain strongly aligned over the entire span, with Cuba and the USSR shifting alignments depending of their political views.}
		\label{fig:embeddings_onu}
	\end{figure}

	Fig. \ref{fig:embeddings_onu} depicts the embeddings of four countries: USA, Israel, Cuba, and the USSR (later, the Russian Federation). We use a similar visualization style as in Fig. \ref{fig:rssi}, with different color palettes used to distinguish among countries, and lighter tones indicating more recent years. Observe how the representations for the USA and Israel remain strongly aligned over the entire time horizon, which is consistent with their longstanding agreement on UN resolution matters. The embedding for the USSR is initially (nearly) orthogonal to the USA and Israel, with Cuba initially showing a greater affinity to the USA/Israel block. This is consistent with Cold War geopolitics of the time. Then, after 1959, Cuba's position shifts to the lower half-plane, becoming more aligned with the USSR. This is expected given Cuba's sharp shift in foreign policy as a result of the Cuban revolution, with its ideology being in agreement with that of the USSR. This polarized scenario remained unchanged until 1991. That year the embedding for the USSR (now the Russian Federation) moves closer to the USA/Israel block, which reflects the politics of the Russian Federation in the aftermath of USSR's dissolution. Cuba remains at an (almost) orthogonal position from the USA/Israel block, with Russia eventually shifting to a middle ground after the mid-2000's.

	
	\section{Concluding Remarks}\label{sec:conclusions}
	
	We developed a gradient-based spectral-embedding framework to estimate latent positions of RDPGs. Relative to prior art our algorithmic approaches offer better representation at a competitive computational cost, and they are more broadly applicable to settings with incomplete, dynamic, and directed network data. We motivated and proposed a novel manifold-constrained formulation to embed directed RDPGs, and developed novel Riemannian GD iterations to estimate interpretable latent nodal positions. The effectiveness of the GRL framework is demonstrated via reproducible experiments with both synthetic and real (wireless network and United Nations voting) data. We made all our codes publicly available.
	
	This work and its current limitations open up several exciting and challenging directions for future research. {For even better scalability, in the near future we plan to migrate our Python implementations to a faster language such as C. Exploring the viability of parallelizing the BCD iterations is another direction in our future research agenda.} With regards to the streaming scenario, it would be of interest to develop lightweight online rules to adaptively determine the embedding dimension. Performing dynamic regret analyses of the online GD methods would be a valuable contribution, since such guarantees in non-convex settings are so far quite rare.

	\bibliographystyle{IEEEtran}
	%
	\bibliography{referencias}
	
	\appendix
	

	%
	%
	
	\subsection{Proof of Proposition \ref{stiefel_tilde_manifold}}
	In order to show that $\mathcal{M}$ is a manifold and to further understand its differential structure, consider the function $F:\reals^{N\times d}_*\mapsto \mathcal{S}_d$ defined as $F(\bbX) = \bbM\circ (\bbX^\top \bbX - \bbI_d)$. Observe that $\mathcal{M}$ is defined as the preimage of zero through $F$, so we will prove that this is a regular value.
	
	
	
	For $\bbzeta \in \reals^{N\times d}$, the derivative of $F$ in $\bbX \in F^{-1}(\bb0_{d\times d})$ along $\bbzeta$ is 
	$DF(\bbX)\bbzeta = \bbM\circ (\bbzeta^\top \bbX + \bbX^\top \bbzeta).$    
	Next we establish that $DF(\bbX)$ is onto. Indeed, let $\bbeta$ be a matrix in the orthogonal complement of the image, i.e., $\bbeta \in ImDF(\bbX)^\perp \subset \mathcal{S}_d$. Then 
	\begin{equation*}
		\langle \bbeta , \bbM\circ (\bbzeta^\top \bbX + \bbX^\top \bbzeta) \rangle = 0, \, \forall \bbzeta \in \reals^{N\times d}.
	\end{equation*}
	Now, since the diagonal of $\bbeta$ is null, we may drop the Hadamard product with $\bbM$ and obtain
	\begin{equation*}
		\langle \bbeta , \bbM\circ (\bbzeta^\top \bbX + \bbX^\top \bbzeta) \rangle = \langle \bbeta ,  \bbzeta^\top \bbX + \bbX^\top \bbzeta \rangle = 0,\, \forall \bbzeta \in \reals^{N\times d}.
	\end{equation*}
	%
	So we have 
	$\tr\left(\bbeta \bbzeta^\top \bbX \right) +  \tr\left(\bbeta \bbX^\top \bbzeta\right)=0$
	and these two summands are equal to each other by virtue of the circular property of the trace operator.
	%
	Hence, we obtain
	$2 \tr\left(\bbeta \bbX^\top \bbzeta\right)=0, \, \forall \, \bbzeta \in \reals^{N\times d},$
	%
	and since this trace vanishes for all $\bbzeta$, we have that $\bbeta\bbX^\top = \bb0_{d\times N}.$ Multiplying by $\bbX$ we obtain $\bbeta\bbX^\top\bbX = \bb0_{d\times d}.$ Becasue $\bbX^\top\bbX$ is diagonal, necessarily $\bbeta = \bb0_{d\times d}$ and therefore $DF(\bbX)$ is onto. The conclusion is that $\mathcal{M}$ is a differential manifold, of dimension $Nd - \frac{d(d-1)}{2}$.
	
	The tangent space at $\bbX$ can be obtained as the kernel of $DF(\bbX)$, so we have
	%
	%
	\begin{equation}\label{eq:tangent}  
		T_{\bbX}\mathcal{M} =  \{\bbzeta  \in \reals^{N\times d}: \bbM \circ \left( \bbzeta^\top \bbX + \bbX^\top\bbzeta\right)=\bb0_{d\times d}  \},   
	\end{equation}
	completing the proof.\hfill$\blacksquare$
	
	
	
	
	
	\subsection{Proof of Lemma  \ref{lemma:charact_normal}}
	Consider a matrix of the form $\bbX\bbLambda$ with $\bbLambda \in \mathcal{S}_d$, and let us show that it is orthogonal to a matrix of the tangent space. Now, observe that $\tr\left((\bbX\bbLambda)^\top \bbzeta \right)=\tr\left(\bbLambda \bbzeta^\top \bbX\right).$
	%
	Therefore,
	%
	\begin{equation*}
		\tr\left((\bbX\bbLambda)^\top\bbzeta \right) = \frac{1}{2}\tr\left(\bbLambda(\bbX^\top \bbzeta+ \bbzeta^\top\bbX)\right)=0.
	\end{equation*}
	The last trace is zero since $\bbLambda \in \mathcal{S}_d$ and $\bbX^\top \bbzeta+ \bbzeta^\top\bbX$ is diagonal, because $\bbzeta \in T_{\bbX}\mathcal{M}$.\hfill$\blacksquare$
	
	As expected, the dimension of the normal space is $\frac{N(N-1)}{2}$, which is the dimension of $\mathcal{S}_d$.
	
	\subsection{Proof of Lemma  \ref{lemma:projec_normal_space} and Proposition \ref{prop:proj_tangent}}
	
	To compute the projection to the normal space, recall 
	some auxiliary matrices defined in Lemma \ref{lemma:projec_normal_space}. 
	Let $\bbX\in\mathcal{M}$. Then $\bbX^\top\bbX$ is diagonal, with positive entries. Define $\bbD = \left(\bbX^\top\bbX\right)^{1/2}$ and let $\bbE = \mathbf{1}_d\mathbf{1}_d^\top \bbD^2 + \bbD^2\mathbf{1}_d\mathbf{1}_d^\top$. These matrices allow us to re-write the operation $\varphi(\bbA) = \bbA\bbD^2 + \bbD^2\bbA$ as $\varphi(\bbA) = \bbA \circ \bbE$. 
	In particular, this allows us to obtain an expression for the inverse operation, which will be needed. Indeed, if $\bbF$ is the matrix with entries $F_{ij}=E_{ij}^{-1}$, then $\left(\bbA\bbD^2 + \bbD^2\bbA\right)\circ \bbF = \bbA$, for all $\bbA \in \reals^{d\times d}$. We can now prove the expression of the projection as follows.
	
	From the characterization of Lemma \ref{lemma:charact_normal}, it is clear that $ \bbX s(2\bbD\bbL) \in T_{\bbX}\mathcal{M}^\perp$, since $s(2\bbD\bbL) \in \mathcal{S}_d$. Let us see that $\bbZ - \bbX s(2\bbD\bbL) \in T_{\bbX}\mathcal{M}$. 
	From \eqref{eq:tangent}, we have to show that
	\begin{equation*}
		\left(\bbZ - \bbX s(2\bbD\bbL)\right)^\top\bbX + \bbX^\top\left(\bbZ - \bbX s(2\bbD\bbL)\right) \,\, \text{is diagonal.}
	\end{equation*}
	Indeed,
	{\small 
		\begin{align*}
			& \left(\bbZ - \bbX s(2\bbD\bbL)\right)^\top\bbX + \bbX^\top\left(\bbZ - \bbX s(2\bbD\bbL)\right) = \\ 
			& \bbZ^\top\bbX - (s(2\bbD\bbL))^\top\bbX^\top\bbX +\bbX^\top\bbZ - \bbX^\top\bbX s(2\bbD\bbL) = \\  
			& \bbZ^\top\bbX +\bbX^\top\bbZ - \frac{1}{2}\left[ 2\bbD\bbL + 2(\bbD\bbL)^\top\right]\bbX^\top\bbX +  \textrm{diag}(2\bbD\bbL)\bbX^\top\bbX -\\ &\frac{1}{2}\bbX^\top\bbX\left[ 2\bbD\bbL + 2(\bbD\bbL)^\top\right] + \bbX^\top\bbX \textrm{diag}(2\bbD\bbL).
		\end{align*}
	}%
	Now, since $\textrm{diag}(2\bbD\bbL)$ and $\bbX^\top\bbX$ are diagonal matrices, they commute, and their product is diagonal. So we can forget those two terms in the expression, and continue with the rest. We will use the expression of $\bbL$ and the fact that $\bbX^\top\bbX = \bbD^2$. Hence,
	{\small 
		\begin{align*}
			& \bbZ^\top\bbX +\bbX^\top\bbZ - \frac{1}{2}\left[ 2\bbD\bbL + 2(\bbD\bbL)^\top\right]\bbX^\top\bbX - \\
			& \frac{1}{2}\bbX^\top\bbX\left[ 2\bbD\bbL + 2(\bbD\bbL)^\top\right] = \\
			& \bbZ^\top\bbX +\bbX^\top\bbZ - \left[ \bbD\bbL\bbD^2 + \bbL^\top\bbD\bbD^2 + \bbD^2\bbD\bbL + \bbD^2\bbL^\top\bbD \right] = \\
			& \bbZ^\top\bbX +\bbX^\top\bbZ - \left[ \bbD\left( \bbL\bbD^2 + \bbD^2\bbL\right) + \left(\bbL^\top\bbD^2 + \bbD^2\bbL^\top \right)\bbD \right] = \\
			& \bbZ^\top\bbX +\bbX^\top\bbZ - \left[ \bbD\left( \bbL \circ \bbE\right) + \left(\bbL^\top \circ \bbE \right)\bbD \right].
		\end{align*}
	}%
	Now, observe that $\bbL\circ \bbE = \left(\left(\bbD^{-1}\bbX^\top \bbZ\right)\circ \bbF\right) \circ \bbE = \bbD^{-1}\bbX^\top \bbZ$, and the same happens with $\bbL^\top$. We end up with
	\begin{equation*}
		\bbZ^\top\bbX +\bbX^\top\bbZ - \left[ \bbD\left( \bbD^{-1}\bbX^\top \bbZ\right) + \left(\bbZ^\top \bbX \bbD^{-1} \right)\bbD \right] = \bb0_{d\times d},
	\end{equation*}
	which in particular is diagonal. Therefore, $\bbZ - \bbX s(2\bbD\bbL) \in T_{\bbX}\mathcal{M}$ and this completes the proof.\hfill $\blacksquare$

	The proof of Proposition \ref{prop:proj_tangent} is straightforward now.
	We have all we need to compute the projection to the tangent space $\pi_\bbX:\reals^{N\times d}\mapsto T_{\bbX}\mathcal{M}$, since $\pi_\bbX(\bbZ) +\pi_\bbX^\perp(\bbZ)= \bbZ.$\hfill $\blacksquare$

	\subsection{Proof of Proposition \ref{prop:retraction}}
	
	Denoting by $\reals^{N\times d}_{fr}$ the set of $N \times d$ full-rank matrices, and by $Supp_1(d)$ the set of upper triangular matrices with ones in the diagonal, let us consider the mapping 
	$$\phi:\mathcal{M}\times Supp_1(d) \mapsto \reals^{N\times d}_{fr}, \, \text{with } \phi(\widetilde{\bbQ},\widetilde{\bbR}) = \widetilde{\bbQ}\widetilde{\bbR}.$$
	
	From the discussion immediately preceding the statement of Proposition \ref{prop:retraction}, we have that $\phi$ is bijective. Furthermore, $\phi$ is smooth since its the restriction of the matrix multiplication to a submanifold. Now, given a full rank matrix $\bbM$, the first component of $\phi^{-1}$ can be obtained as the result of a modified Gram-Schmidt process, which is is differentiable. The second component can then be obtained as $\widetilde{\bbR} = \widetilde{\bbQ}^{-1}\bbM$, and therefore it is also differentiable. It follows that $\phi$ is a diffeomorfism.
	
	We also have that $\phi(\widetilde{\bbQ},\bbI_d) = \widetilde{\bbQ}$. Following \cite[Prop. 4.1.2]{absil} we have that the projection onto the first component of $\phi^{-1}$ is a retraction, which is exactly the $\widetilde{qf}$ mapping defined in Proposition \ref{prop:retraction}.\hfill$\blacksquare$ 
	
	{
		\subsection{Robustness to initialization}\label{sec:app_robust}
		Since the objective functions we optimize are non-convex and convergence guarantees provided are to stationary solutions, it is prudent to study the algorithms' sensitivity to the initialization. As discussed in Section \ref{sec:grad_desc}, except for specific problematic initializations corresponding to sub-optimal stationary points, in our experience all algorithms converge to the optimum when they are initialized at random. The following experiment illustrates this desirable property, in particular for the Riemannian GD (i.e.,\ Algorithm \ref{algo:mani_grad_descent}) method developed to embed digraphs. Similar results are obtained for the other algorithms, but not included here to avoid repetition. 
		
		\begin{figure}[t]
			\centering
			\includegraphics[width=0.95\linewidth]{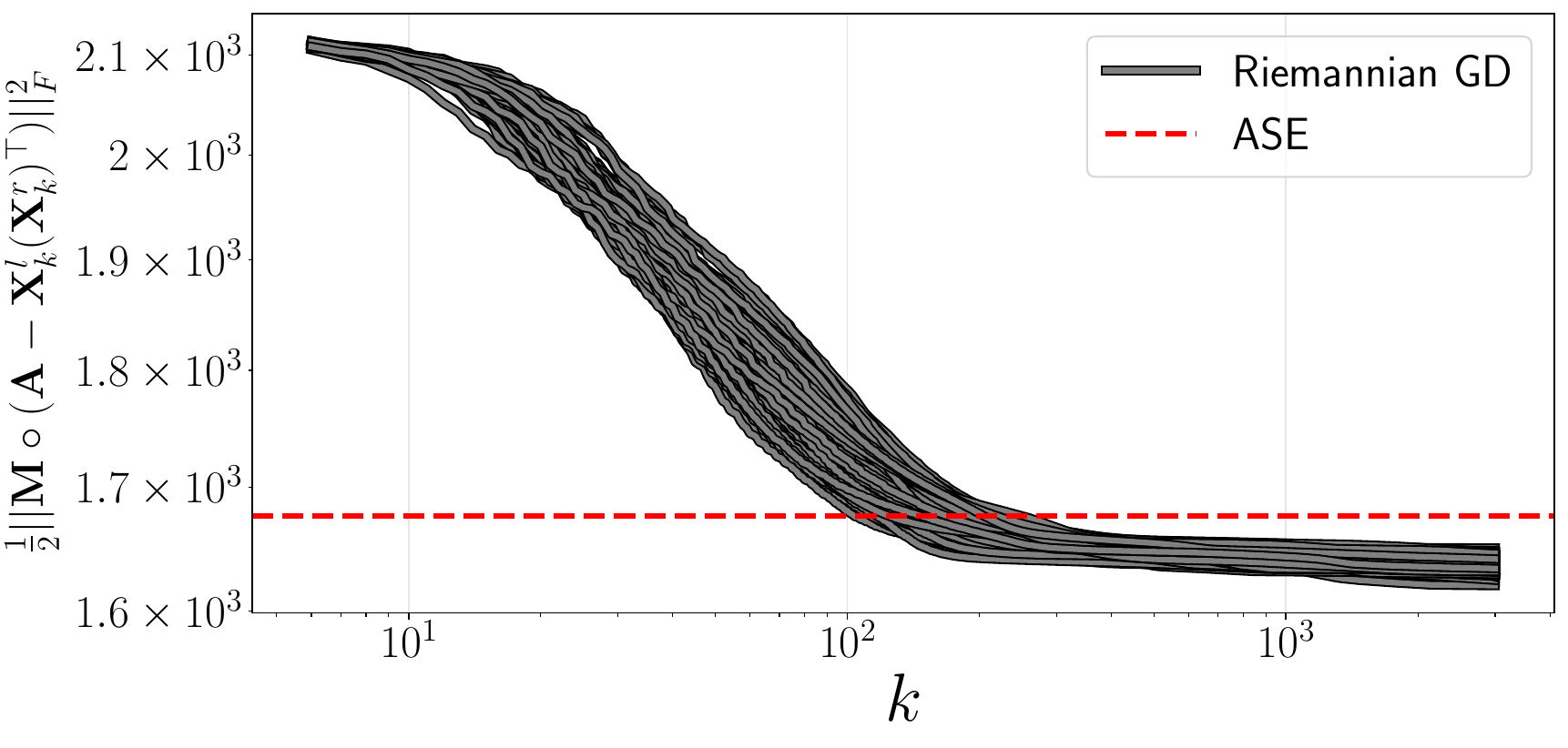}
			\caption{{The evolution of $f(\bbX^l_k,\bbX^r_k)=\frac{1}{2}\left\|\bbM \circ (\bbA-\bbX_k^l(\bbX_k^r)^\top)\right\|_F^2$ using Algorithm \ref{algo:mani_grad_descent} to embed an LFR graph, starting from 75 different random initializations. The first 5 iterations are omitted for clarity. Note how the algorithm systematically produces estimates of the embeddings with a lower cost than ASE, and marginal variability regardless of the initialization.}}
			\label{fig:robust}
		\end{figure}
		
		We consider a Lancichinetti–Fortunato–Radicchi (LFR)~\cite{lrs2008benchmark} benchmark graph with $N=1000$ nodes, randomly initialize Algorithm \ref{algo:mani_grad_descent} and plot the evolution of the cost function $f(\bbX^l_k,\bbX^r_k)$ in \eqref{eq:dase_mask_constrain}. The LFR model is a widely adopted benchmark that produces graphs with properties observed in real-world networks, particularly in terms of the resulting degree distribution and community sizes. Here, the LFR graph was generated using parameters $\tau_1=3$ and $\tau_2=2$ as exponents of the power law distributions for the degree and community size, respectively, and mixing parameter $\mu=0.1$. The resulting graph has $16$ communities, the larger one with $142$ nodes, and the smaller one with $30$ nodes. The largest hub has $157$ neighbors, and there are several nodes with degree $2$.
		Fig. \ref{fig:robust} shows the results over 75 randomly initialized runs where $d=16$, and also the cost function value $1676.49$ obtained by ASE. Regardless of the initialization, the limiting objective values obtained via Riemannian GD exhibit marginal variability ($\textrm{mean}=1635.66$, $\textrm{std}=6.52$) and always outperform ASE. In terms of timing, embedding each of these LFR graphs with $N=1000$ nodes takes 40 seconds on average.}
	
\end{document}